\documentclass{article}

    \PassOptionsToPackage{numbers, compress}{natbib}
 \usepackage[preprint]{neurips_2026}

\usepackage[utf8]{inputenc} 
\usepackage[T1]{fontenc}    
\usepackage{hyperref}       
\usepackage{url}            
\usepackage{booktabs}       
\usepackage{amsfonts}       
\usepackage{nicefrac}       
\usepackage{microtype}      
\usepackage[table]{xcolor}  
\usepackage{graphicx}
\usepackage{multirow}
\usepackage{amsmath}
\usepackage{placeins}
\usepackage{capt-of}
\usepackage{wrapfig}
\usepackage{pifont}

\newcommand{\cmark}{\ding{51}}
\newcommand{\xmark}{\ding{55}}

\title{Georeferenced Cross-View 3D Reconstruction from Ground and Satellite Images}

\author{%
  \begin{tabular}{cc}
    David G. Shatwell & Mubarak Shah \\
    \texttt{david.shatwell@ucf.edu} & \texttt{shah@crcv.ucf.edu}
  \end{tabular}\\
  {} \\
  \normalfont Institute of Artificial Intelligence, University of Central Florida
}

\begin{document}

\maketitle

\begin{abstract}
Georeferenced 3D reconstruction from ground images requires recovering not only the local scene geometry, but also localizing every camera center and 3D point in a geographic coordinate frame. Existing cross-view geo-localization methods are able to estimate the GPS coordinate of a single ground camera, but do not reconstruct dense 3D geometry, while recent feed-forward multi-view 3D reconstruction models predict point maps and camera poses only in a local coordinate frame. We introduce X-VGGT, a feed-forward cross-view geometry transformer that unifies these tasks by jointly processing a set of ground-level images and a georeferenced satellite image to predict ground-view camera poses, dense 3D point maps, and a scene-level transform parameterized by gravity, scale, yaw and translation. This transform projects the local reconstruction into the satellite plane, resulting in the geo-registration of all ground images with the georeferenced satellite image. We further propose an evaluation protocol for this joint task and show that, across multiple cross-view datasets, X-VGGT produces accurate georeferenced reconstructions and outperforms prior cross-view geo-localization and multi-view 3D reconstruction models. Project page: \url{https://dshatwell23.github.io/xvggt-page/}.
\end{abstract}

\section{Introduction}

Ground-level images are easy to collect and can be used to reconstruct detailed 3D models; however, image-based reconstruction methods typically recover geometry only up to an unknown global scale. In recent years, high-resolution satellite images have become widely available across many regions of the world. Since these images are often georeferenced, their image coordinates can be mapped to real-world geographic locations, providing GPS information for each pixel. This makes them an ideal complementary source of information for image-based 3D reconstruction, as they can provide not only a metric reference for resolving scale ambiguity, but also a global coordinate frame for localizing the reconstructed cameras and 3D points.

However, using satellite images as a geographic reference requires solving a difficult cross-view registration problem: ground-level and satellite images observe the same scene from drastically different viewpoints, with large changes in appearance, scale, and visible structure. A model that can overcome this gap would enable practical applications in autonomous driving and robotics in GPS-denied areas, disaster response from localized post-event imagery, and infrastructure inspection, where observed assets and defects could be placed directly on a map.

Prior work addresses important parts of this problem, but not the full setting. Cross-view geo-localization methods are trained to estimate ground-view camera locations within a satellite image \citep{shi2022beyond,wang2023fine,song2023learning,shi2024weakly,wollam2026towards,xia2025fg,lehyeh2026geoflow}, but they are not designed to localize individual image pixels. This capability is known as geo-registration, and requires additional 3D understanding about the scene geometry. Recent feed-forward 3D reconstruction models are able recover camera poses and 3D point maps from sparse views \citep{wang2024dust3r,leroy2024grounding,wang2025vggt,lin2025depth,wang2025pi}, but they typically generate predictions in a local coordinate frame and are known to fail under extreme viewpoint changes.

\begin{figure}[t]
  \centering
  \includegraphics[width=\textwidth]{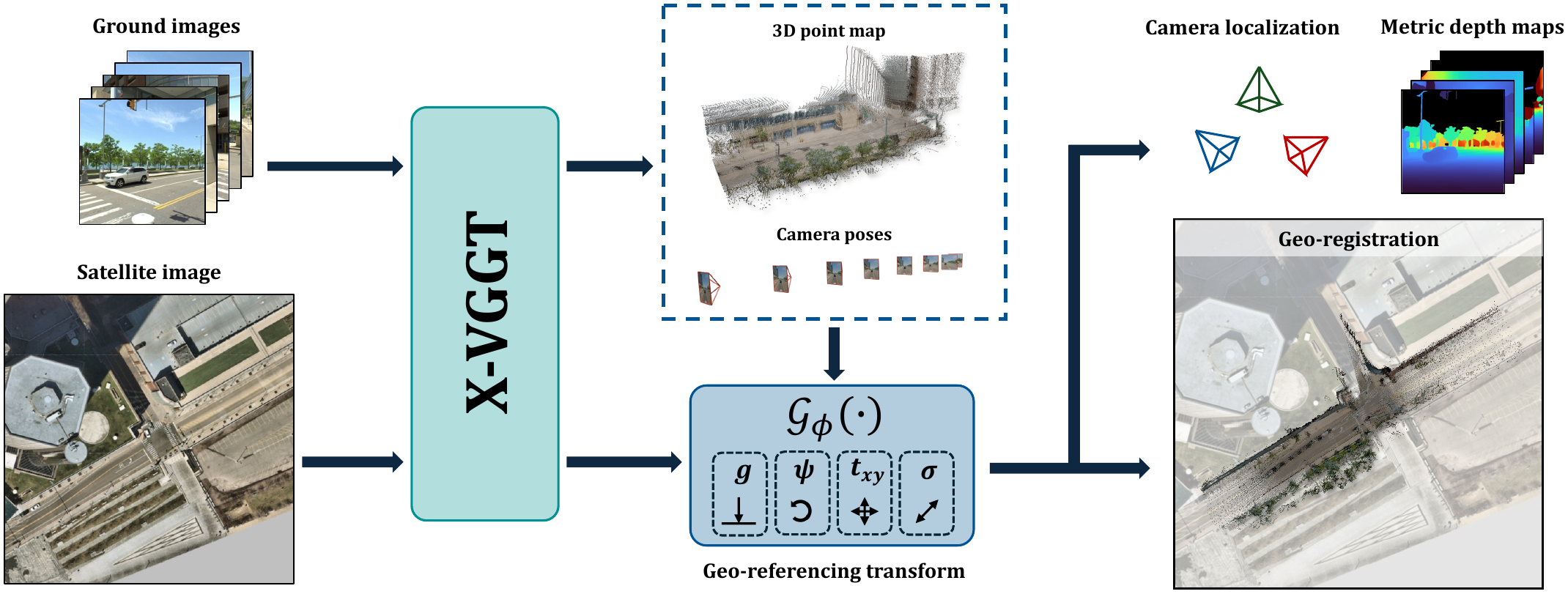}
  \caption{\textbf{Georeferenced cross-view 3D reconstruction.}
  Given a set of ground-level images and a georeferenced satellite view, \textbf{X-VGGT} predicts camera poses, 3D point-maps, and a georeferencing transform $\mathcal{G}_\phi(\cdot)$. By projecting the 3D reconstruction into the satellite plane, our model can geo-register every pixel in the ground images, localize their cameras, and generate metric depth maps.}
  \label{fig:teaser}
\end{figure}

Georeferenced cross-view 3D reconstruction lies at the intersection of these two tasks: using both ground and satellite images, the goal is to generate a metric 3D reconstruction where all ground-level image pixels are localized in a geographic coordinate frame. To address this problem, we introduce \textbf{X-VGGT}, a Cross(\textbf{X})-\textbf{V}iew \textbf{G}eometry \textbf{G}rounded \textbf{T}ransformer for georeferenced cross-view 3D reconstruction. Given a set of ground images and a georeferenced satellite image, X-VGGT predicts ground-view camera extrinsics, intrinsics, metric dense 3D point maps, and a scene-level satellite alignment transform used to geo-localize image pixels and camera centers, as shown in Figure~\ref{fig:teaser}. The satellite image is processed jointly with the ground images through a multi-view transformer, allowing them to exchange information. Then, a dedicated Satellite Alignment Head predicts a ground-to-satellite coordinate transform parameterized by gravity, yaw, scale, and translation, which projects the reconstructed camera centers and dense point maps into the georeferenced satellite plane, thus enabling dense pixel-level geo-registration.

In summary, our key contributions are as follows:
\begin{itemize}
    \item We propose X-VGGT, a feed-forward cross-view 3D reconstruction model
    that jointly uses ground images and a georeferenced satellite image to
    predict cameras, metric 3D point maps, and a satellite-ground alignment transform that geo-localizes all ground image pixels.
    \item We enable georeferenced 3D reconstruction from flexible sets of ground-view
    inputs: X-VGGT grounds camera centers and dense point maps in the satellite
    coordinate frame while supporting ordered sequences or unordered sets of
    limited-FOV images.
    \item We introduce a Satellite Alignment Head and georeferencing losses that directly supervise the ground-to-satellite transformation, projected camera centers, and pixel-wise geo-registration.
    \item We introduce new training and evaluation protocols for cross-view 3D reconstruction, using ordered and unordered sets of frames from unseen locations to evaluate the generalization of X-VGGT and competing baselines.
\end{itemize}

\section{Related Work}

\textbf{Geo-localization} aims to estimate where an image was captured by predicting the camera GPS coordinates using only visual information. Existing approaches are broadly grouped into classification, retrieval, or registration. Classification methods discretize the Earth into geographic cells and use the centroid of the predicted geocell as the estimated GPS coordinate \citep{weyand2016planet,vo2017revisiting,seo2018cplanet,muller2018geolocation,pramanick2022world,clark2023we}. Retrieval methods embed the query into a high-dimensional feature space and compare it against a geotagged gallery, formed either by encoding GPS coordinates with a location encoder \citep{vivanco2023geoclip,klemmer2025satclip,shatwell2025gt,astruc2026unigeoclip,haas2024pigeon,Astruc_2024_CVPR,shatwell2026tiger}, or by using images with known locations as proxies.

Image-retrieval methods can be divided into same-view and cross-view settings. Same-view visual place recognition (VPR) embeds query and gallery images with a shared image encoder and ranks the gallery by feature similarity \citep{torii2013visual,arandjelovic2016netvlad,berton2022rethinking,ali2022gsv,ali2023mixvpr,keetha2023anyloc,berton2023eigenplaces,tzachor2024effovpr}. Cross-view geo-localization instead matches ground-level queries against satellite images, using either shared or view-specific encoders. Prior work has improved this framework with ground-view warping, hard negative mining, re-ranking, stronger backbones, and auxiliary objectives \citep{zhu2022transgeo,deuser2023sample4geo,li2024unleashing,zhu2021vigor,ye2024cross}. Cross-view retrieval has also been extended to videos using multi-stage pipelines and temporal aggregation \citep{zhang2023cross,vyas2022gama,pillai2024garet}; however, these methods typically assume ordered image sequences and are less suited to unstructured multi-view scenes.

A key limitation of retrieval-based geo-localization is that predictions remain tied to the gallery discretization: the estimated GPS coordinates represent the center pixel of a query, rather than pixel-wise GPS locations of the query image. Recent cross-view localization methods address this by registering the ground view within a georeferenced satellite image, typically through pose search, BEV transformations, dense correspondence fields, or fine-grained cross-view matching \citep{shi2022beyond,wollam2026towards,wang2023fine,song2023learning,wang2025bevsplat,lehyeh2026geoflow,shi2024weakly,xia2025fg,xia2025loc}. However, they remain primarily 2D pose-estimation systems: most predict a single rotation and translation within a fixed satellite area from a single ground image or panorama, without reconstructing dense 3D structure or placing multiple cameras and point maps in a shared georeferenced frame. This motivates moving beyond local 2D registration toward georeferenced multi-view 3D reconstruction.

\textbf{Multi-view 3D reconstruction} recovers camera geometry and dense scene structure rather than only a 2D pose. Recent feed-forward geometry models directly predict this structure from images: DUSt3R \citep{wang2024dust3r} and MASt3R \citep{leroy2024grounding} estimate point maps and correspondences from uncalibrated image pairs, and several methods extend this paradigm to larger or more structured inputs through multi-view prediction, memory, recurrent state, relative-pose aggregation, or video geometry \citep{Cabon_2025_CVPR,wang2025cut3r,Dong_2025_CVPR,ICLR2025_ce1710ff}. In parallel, learning-based SfM and general multi-view reconstruction models are increasingly adapted to predict cameras, depth, point maps, visual appearance, or metric 3D from sparse or dense scenes \citep{wang2024vggsfm,wang2025vggt,lin2025depth,wang2025pi,Yang_2025_CVPR,Zhang_2025_CVPR,keetha2026mapanything}.

However, these models are not designed for georeferenced ground-satellite reconstruction. They typically recover geometry in a local coordinate frame with arbitrary scale and rely on sufficient visual overlap, whereas nadir satellite and street-level images differ substantially in viewpoint, appearance, and visibility. Recent methods partially address this gap by adapting 3D reconstruction models to drone-ground pairs~\citep{vuong2025aerialmegadepth}, drone-satellite pairs~\cite{liu2026pixel}, ground images and top-down floor plans~\citep{huang2025c3po}, or satellite, aerial, and ground views~\citep{wang2026sky2ground}. While promising, these approaches do not directly solve ground-satellite registration with real orthorectified satellite imagery: they either rely on drone viewpoints, use non-standard top-down inputs, or are trained with synthetic perspective-projection aerial images.

While extreme-view methods make important progress, they do not fully address the setting targeted by our method: reconstructing arbitrary ordered or unordered sets of ground views directly in the coordinate frame of a georeferenced orthorectified satellite image, without using any auxiliary modalities. In this work, our aim is to connect cross-view localization with multi-view 3D reconstruction, using satellite images to geo-register all cameras and pixels in a geographic reference frame.

\section{Method}

\begin{figure}[t]
  \centering
  \includegraphics[width=\textwidth]{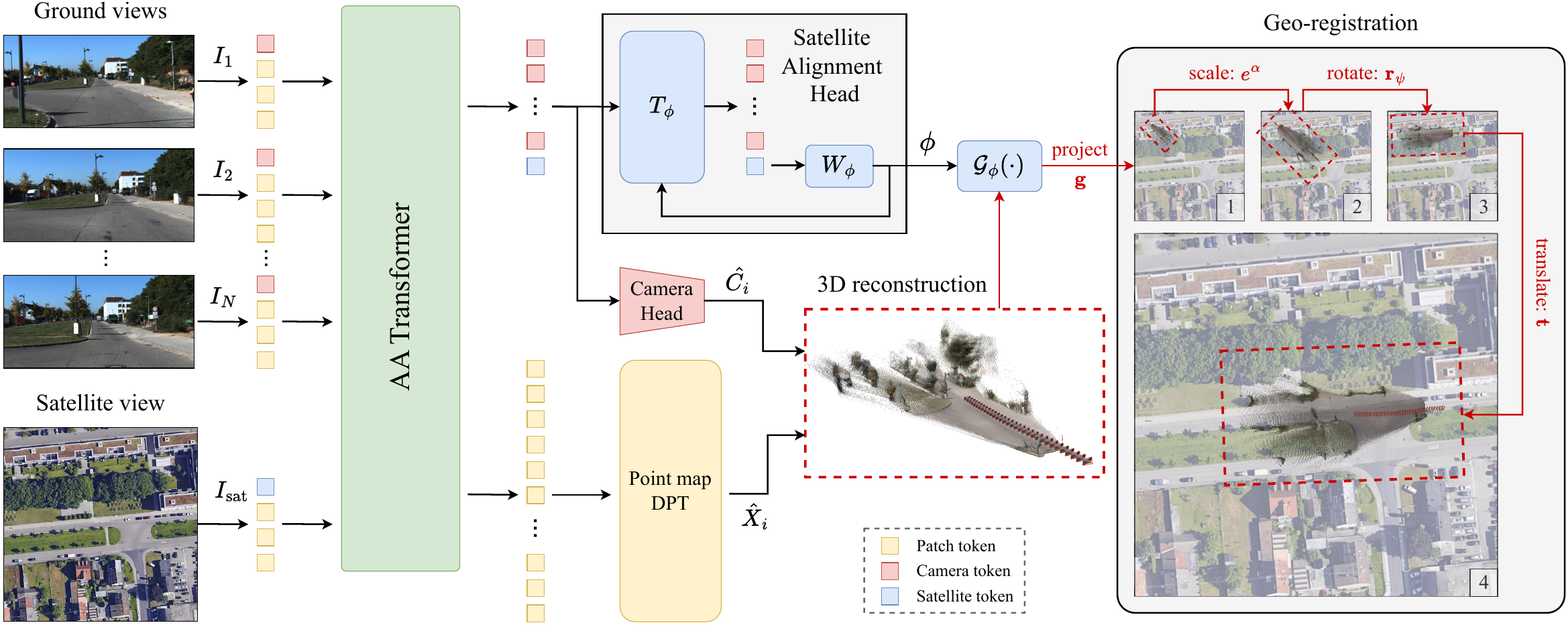}
\caption{\textbf{Overview of X-VGGT.}
The model jointly processes \(N\) ground-level images and a georeferenced satellite image with an Alternating Attention (AA) Transformer. Ground-view patch tokens are decoded by the camera and DPT heads to predict camera poses \(\hat{C}_i\) and dense point maps \(\hat{X}_i\), while the satellite token is decoded by the Satellite Alignment Head (SAH) to estimate the transformation \(\mathcal{G}_\phi\) from the local 3D reconstruction to the satellite coordinate frame. \(\mathcal{G}_\phi\) is parameterized by a gravity vector \(\mathbf{g}\) for ground-plane projection, a scalar \(\alpha\) for metric conversion, and a planar rotation \(\psi\) and translation \(\mathbf{t}\) that align the projected points with the satellite image.}
\label{fig:architecture}
\end{figure}

Given a set of ground images and a georeferenced satellite image, X-VGGT predicts ground-view camera poses, dense 3D point maps, and a satellite-to-ground transform that is used to geo-register all cameras and image pixels. In this section, we first formulate the georeferenced 3D reconstruction problem (Sec.~\ref{sec:geo3d}), then describe the architecture (Sec.~\ref{sec:xvggt}), training objectives (Sec.~\ref{sec:losses}), and training strategy (Sec.~\ref{sec:training}). Figure \ref{fig:architecture} provides an overview of our method.

\subsection{Problem Formulation}
\label{sec:geo3d}

Let \(I_{\mathrm{sat}}\) denote an orthorectified georeferenced satellite
image, where image coordinates can be mapped to latitude-longitude pairs, and let
\(\{I_i\}_{i=1}^{N}\) denote \(N\) ground-level images captured within the
satellite area but with unknown geo-location. The ground inputs may be
an ordered sequence, an unordered set, or perspective views extracted from a
panorama. X-VGGT predicts camera extrinsics
\(\hat{E}_i \in \mathbb{R}^{3 \times 4}\), intrinsics
\(\hat{K}_i \in \mathbb{R}^{3 \times 3}\), dense 3D point maps
\(\hat{X}_i \in \mathbb{R}^{H \times W \times 3}\), and a georeferencing transform
\(\mathcal{G}_\phi: \mathbb{R}^3 \rightarrow [0,1]^2\) used to warp pixels from the local ground-views into the satellite image plane, allowing dense point maps and camera centers to be projected and geo-localized.

The georeferencing map is a single scene-level transform. Rather than
predicting each camera location independently, X-VGGT predicts one geometric
map that aligns the full local reconstruction with the satellite plane. We
parameterize this map by
\[
    \phi = (\mathbf{g}, \mathbf{r}_\psi, \alpha, \mathbf{t}),
    \qquad
    \mathbf{r}_\psi = (\cos\psi,\sin\psi),
\]
where \(\mathbf{g}\in\mathbb{R}^3\) is a unit gravity vector in the local
reconstruction frame, \(\mathbf{r}_\psi \in \mathbb{R}^2\) is a unit yaw vector
that rotates the horizontal basis, \(\alpha=\log s\) is the logarithmic scaling factor,
and \(\mathbf{t}\in\mathbb{R}^2\) is a translation vector in normalized satellite
coordinates.

The gravity direction defines the normal of the satellite-aligned plane. To
obtain in-plane axes, we project a canonical reference direction onto the plane
orthogonal to \(\mathbf{g}\). Let \(\mathbf{a}\) be the normalized projection
of the \(x\)-axis onto this plane and let \(\mathbf{b}=\mathbf{g}\times
\mathbf{a}\); the yaw vector is used to rotate the basis as follows:
\[
    \mathbf{x}_\phi = \cos\psi\,\mathbf{a} + \sin\psi\,\mathbf{b},
    \qquad
    \mathbf{y}_\phi = -\sin\psi\,\mathbf{a} + \cos\psi\,\mathbf{b}.
\]
For any reconstructed point \(\mathbf{X}\), the satellite projection is thus given by
\[
    \mathcal{G}_\phi(\mathbf{X})
    =
    \exp(\alpha)
    \left[
    \begin{array}{c}
    \mathbf{x}_\phi^\top \mathbf{X} \\
    \mathbf{y}_\phi^\top \mathbf{X}
    \end{array}
    \right]
    + \mathbf{t}.
\]
This orthographic projection maps camera centers and dense points from the local
reconstruction frame into normalized satellite coordinates, with the vertical
axis pointing upward. For a satellite crop spanning \(\rho_{\mathrm{sat}}\) meters,
\(\rho_{\mathrm{sat}}e^\alpha\) gives the metric scale in meters per reconstruction unit.

\subsection{Cross-View Geometry Grounded Transformer}
\label{sec:xvggt}

X-VGGT jointly processes the ground and satellite images with a
multi-view transformer using alternating attention layers. The satellite image is
appended as the final view, giving the input tuple
\((I_1,\ldots,I_N,I_{\mathrm{sat}})\) with \(M=N+1\) total views. A shared
visual encoder, initialized from DINOv2 \citep{oquab2023dinov2}, converts each
image into \(P\) patch tokens ($\mathbf{p}_i$). Camera ($\mathbf{c}_i$) and register ($\mathbf{z}^{\mathrm{reg}}_{i,j}$) tokens are appended to every
view. All tokens have embedding dimension \(D=1024\):
\[
    Z^0_i =
    [\,\mathbf{c}_i,\mathbf{z}^{\mathrm{reg}}_{i,1},\ldots,\mathbf{z}^{\mathrm{reg}}_{i,N_r},
    \mathbf{p}_{i,1},\ldots,\mathbf{p}_{i,P}\,]
    \in \mathbb{R}^{(1+N_r+P)\times D},
    \qquad i=1,\ldots,M .
\]
We use three sets of learnable camera and register tokens: one for the reference ground view $I_1$, one shared by the remaining ground views $\{I_i\}_2^N$, and one for the satellite view $I_\mathrm{sat}$. A shared image encoder produces the patch tokens for all views.

Following the alternating-attention design of VGGT \citep{wang2025vggt}, each
block alternates between frame attention, which updates tokens within each view,
and global attention, which updates tokens across the concatenated set of views:
\[
    \bar{Z}^{\ell}_i = F_\ell(Z^{\ell-1}_i),
    \qquad
    Z^{\ell} =
    G_\ell([\,\bar{Z}^{\ell}_1;\ldots;\bar{Z}^{\ell}_M\,]).
\]
This preserves view-specific image evidence while repeatedly exchanging
information across the ground and satellite views.

The satellite image has a different geometric role from a ground camera. Since
the satellite image is orthorectified and used as a nadir map, X-VGGT does
not predict a perspective camera pose, finite camera center, depth map, or point
map. Instead, the satellite tokens condition the ground
3D reconstruction through attention and are read by a dedicated Satellite
Alignment Head.

\textbf{Satellite Alignment Head.}
The Satellite Alignment Head (SAH) predicts scene-level georeferencing
parameters \(\phi\). Let
\(Y\in\mathbb{R}^{M\times 2D}\) denote the camera-token features from all
\(M\) views, including satellite. Each row concatenates the
\(D\)-dimensional camera-token embeddings from the final frame-attention and
global-attention blocks, yielding \(2D=2048\) features per view. At each refinement step \(k\), the head
embeds the previous estimate \(\tilde{\phi}^{k-1}\), mixes the \(M\) camera
tokens with a lightweight transformer \(T_\phi\), and reads out only the final
satellite token:
\[
    U^k = T_\phi(Y;\tilde{\phi}^{k-1}),
    \qquad
    \tilde{\phi}^{k}
    =
    \tilde{\phi}^{k-1}
    +
    W_\phi(U^k_{\mathrm{sat}}),
    \qquad k=1,\ldots,4 .
\]
For the first step, \(\tilde{\phi}^{0}\) is represented by a learned empty
georeferencing token. The final activated estimate
\(\phi=A(\tilde{\phi}^{4})\) normalizes \(\mathbf{g}\) and
\(\mathbf{r}_\psi\) to unit length while leaving \(\alpha\) and
\(\mathbf{t}\) unconstrained. Reading only from the satellite token reflects
the role of \(\phi\): it describes the alignment between the reconstructed
ground scene and the satellite frame.

\textbf{3D Geometry Heads.}
The ground-view geometry heads follow the VGGT design. The camera head uses an
iterative transformer over the \(N\) ground camera tokens and predicts a
pose encoding in \(\mathbb{R}^{N\times 9}\), consisting of 3D translation, a
4D quaternion rotation, and 2D field-of-view parameters for each ground view.
The point head is a dense prediction transformer (DPT)~\cite{ranftl2021vision} that converts ground patch
tokens into dense point maps and confidence maps,
\(\hat{X}\in\mathbb{R}^{N\times H\times W\times 3}\). Both of these heads receive only
the ground-view tokens as input; the satellite patch tokens are not directly ingested by any downstream head, and instead are used as context during the joint encoding stage.

\subsection{Geo-registration}

The camera head predicts a world-to-camera transform
\(\hat{E}_i=[\hat{R}_i\mid\hat{\mathbf{t}}_i]\) for each ground view in the
local reconstruction frame. We recover the camera center as
\(\hat{\mathbf{C}}_i = -\hat{R}_i^\top\hat{\mathbf{t}}_i\) and project both dense point maps and camera centers into the satellite image:
\[
    \hat{\mathbf{u}}_i(p) =
    \mathcal{G}_\phi(\hat{\mathbf{X}}_i(p)),
    \qquad
    \hat{\mathbf{u}}^{\mathrm{cam}}_i =
    \mathcal{G}_\phi(\hat{\mathbf{C}}_i),
\]
where \(p\) indexes image pixels. The projected camera centers provide a
fine-grained localization signal, while the projected dense points measure
whether the reconstructed geometry is correctly grounded in the satellite
frame.

\subsection{Training Objectives}
\label{sec:losses}

X-VGGT is trained with a multi-task objective that combines standard 3D
reconstruction losses with new georeferencing losses. The georeferencing losses
supervise the scene-level map parameters and their induced satellite
projections. Let \(\phi^*\) denote the ground-truth georeferencing parameters
computed in the normalized reconstruction frame. For each Satellite Alignment Head (SAH) refinement stage
\(k\), we supervise the activated estimate
\(\phi^k=(\mathbf{g}^k,\mathbf{r}_\psi^k,\alpha^k,\mathbf{t}^k)\) with
\[
    \mathcal{L}_{\phi}
    =
    \frac{1}{K_\phi}
    \sum_{k=1}^{K_\phi}
    \gamma^{K_\phi-k}
    \left[
    1-\langle \mathbf{g}^k,\mathbf{g}^*\rangle
    +
    1-\langle \mathbf{r}_\psi^k,\mathbf{r}_\psi^*\rangle
    +
    |\alpha^k-\alpha^*|
    +
    \|\mathbf{t}^k-\mathbf{t}^*\|_1
    \right],
\]
where \(K_\phi=4\) and \(\gamma=0.6\). The direction terms supervise the unit
gravity and yaw vectors, while the remaining terms supervise scale and
translation.

We also supervise the satellite projections induced by the final
georeferencing transform. Specifically, we compare both the dense
projected point maps and the projected camera centers against their
ground-truth normalized satellite coordinates. The dense loss is computed
over valid pixels \(\Omega_{\mathrm{pts}}\), while the camera-center loss is
computed over the \(N\) ground-view cameras:
\[
\begin{aligned}
    \mathcal{L}_{\mathrm{dense}}
    &=
    \frac{1}{|\Omega_{\mathrm{pts}}|}
    \sum_{(i,p)\in\Omega_{\mathrm{pts}}}
    \left\|
    \mathcal{G}_\phi(\hat{\mathbf{X}}_i(p))
    -
    \mathbf{u}_i(p)
    \right\|_1,
    &
    \mathcal{L}_{\mathrm{center}}
    &=
    \frac{1}{N}
    \sum_{i=1}^{N}
    \left\|
    \mathcal{G}_\phi(\hat{\mathbf{C}}_i)
    -
    \mathbf{u}^{\mathrm{cam}}_i
    \right\|_1 .
\end{aligned}
\]

The remaining losses follow standard feed-forward 3D reconstruction training.
The camera loss \(\mathcal{L}_{\mathrm{cam}}\) applies an L1 loss to the
predicted pose encoding, decomposed into translation, quaternion rotation, and
field-of-view terms. The point-map loss \(\mathcal{L}_{\mathrm{pts}}\)
combines confidence-weighted point regression and a
normal consistency term over valid pixels. The final objective is given by
\[
    \mathcal{L}
    =
    \lambda_{\mathrm{cam}} \mathcal{L}_{\mathrm{cam}}
    +
    \lambda_{\mathrm{pts}} \mathcal{L}_{\mathrm{pts}}
    +
    \lambda_{\phi} \mathcal{L}_{\phi}
    +
    \lambda_{\mathrm{center}} \mathcal{L}_{\mathrm{center}}
    +
    \lambda_{\mathrm{dense}} \mathcal{L}_{\mathrm{dense}} .
\]
All georeferencing and point-map terms have unit weight except for the camera loss, which uses a weight of \(5.0\).

\subsection{Training Strategy}
\label{sec:training}

\textbf{Supervision signals.}
Each training sample contains calibrated ground images, depth-derived point
maps, ground camera poses in an East-North-Up (ENU) reference frame, and an orthorectified
satellite image with known spatial resolution. During training, the data-loaders project the ENU
camera centers into the satellite image to obtain normalized camera-center
targets \(\mathbf{u}^{\mathrm{cam}}_i\in[0,1]^2\). After the standard
normalization of camera poses and point maps into the local reconstruction
frame, we recompute all georeferencing targets in the same coordinate system as
the model predictions. Let \(E_i=[R_i\mid \mathbf{t}_i]\) be the normalized
world-to-camera pose, \(\mathbf{C}_i=-R_i^\top\mathbf{t}_i\) the camera center,
and \(\mathbf{e}_z=(0,0,1)^\top\) the ENU up direction. If \(R^{\mathrm{enu}}_0\)
is the reference-camera rotation before normalization and \(\bar{R}_0\) is the
reference-camera rotation after normalization, the ground-truth gravity in the
normalized reconstruction frame is \(\mathbf{g}^* = \mathrm{norm}(\bar{R}_0^\top R^{\mathrm{enu}}_0\mathbf{e}_z)\).

Given \(\mathbf{g}^*\), we solve for the yaw, scale, and translation that best
align the normalized ground camera centers with their satellite coordinates:
\[
    (\mathbf{r}^*_\psi,\alpha^*,\mathbf{t}^*)
    =
    \arg\min_{\mathbf{r}_\psi,\alpha,\mathbf{t}}
    \sum_{i=1}^N
    \left\|
    \mathcal{G}_{(\mathbf{g}^*,\mathbf{r}_\psi,\alpha,\mathbf{t})}
    (\mathbf{C}_i)
    -
    \mathbf{u}^{\mathrm{cam}}_i
    \right\|_2^2 ,
\]
Dense satellite
targets are then generated by projecting each valid depth-derived 3D point
through the fitted transform \(\mathbf{u}_i(p)=\mathcal{G}_{\phi^*}(\mathbf{X}_i(p))\), for \(p\in\Omega_{\mathrm{pts}}\).

\textbf{Training datasets.}
We train our model with KITTI \citep{geiger2013vision,shi2022beyond} and a modified version of Aerial-MegaDepth \citep{vuong2025aerialmegadepth}, which we call Sat-MegaDepth. KITTI
provides street-level sequences with calibrated camera poses, LiDAR-derived depth maps, and georeferenced satellite views. We construct Sat-MegaDepth from Aerial-MegaDepth
\citep{vuong2025aerialmegadepth} by adding orthorectified satellite images, which are not provided by the original dataset. Unlike KITTI, MegaDepth-style images are scattered across a scene rather than
arranged as a vehicle trajectory. Training with both datasets encourages X-VGGT to handle sequential
or unordered ground-view sets. Additional details on training data, satellite crops, augmentations, dataset construction, and optimization are provided in Appendix~\ref{app:method_details}.

\section{Experiments}

We evaluate X-VGGT on three georeferenced reconstruction tasks: camera geo-localization, pixel-wise geo-registration, and local 3D reconstruction. We first describe the datasets, scene construction, baselines, and metrics used throughout the section, then report quantitative results and ablations.

\subsection{Evaluation Datasets}

We evaluate X-VGGT on four cross-view datasets spanning different locations, sensors, and scene layouts. KITTI~\citep{geiger2013vision,shi2022beyond} provides driving sequences with LiDAR-derived depth and georeferenced satellite images; although KITTI is used during training, evaluation is performed on the geographically disjoint cross-area Test-2 split. We use the augmented KITTI release of \citet{shi2022beyond}, which supplements the original dataset with Google Maps satellite imagery at zoom level 18, corresponding to a spatial resolution of 0.20~m/pixel. Mapillary Metropolis provides synchronized multi-directional driving images, camera poses, MVS-derived depth, and satellite images, and is used for zero-shot evaluation. We further evaluate camera geo-localization on WRIVA~\citep{iarpa2024wriva} and ACC-NVS1~\citep{sugg2025accnvs1}, which contain unseen locations with calibrated ground images and GPS annotations but no ground-truth depth maps. Additional dataset and satellite-image construction details are provided in Appendix~\ref{app:evaluation_details}.

\subsection{Scene Generation}

We construct evaluation scenes according to the specific scene layout of each dataset: for KITTI and ACC-NVS1, we use sequences of consecutive video frames; for Mapillary Metropolis, we use consecutive multi-directional camera groups pointing along the front, back, left, and right sides of the vehicle frame; and for WRIVA, we create unordered image sets with similar camera poses. Each scene consists of \(N\) ground images and a \(150\mathrm{m}\times150\mathrm{m}\) satellite view sampled around the selected ground cameras. We create satellite crops such that the centroid of the ground cameras is randomly placed within the central \(100\mathrm{m}\times100\mathrm{m}\) region of the satellite image. We sample up to \(1024\) scenes per dataset for the main evaluation and use larger scenes with up to \(96\) ground images for the ablations. Additional scene construction details are provided in Appendix~\ref{app:evaluation_details}.

\subsection{Baselines and Evaluation Protocols}
\begin{wraptable}{r}{0.27\textwidth}
\vspace{-1.0\baselineskip}
\centering
\scriptsize
\setlength{\abovecaptionskip}{0pt}
\setlength{\belowcaptionskip}{4pt}
\caption{Native prediction support for each method.}
\label{tab:baseline_prediction_support}
\setlength{\tabcolsep}{2.5pt}
\renewcommand{\arraystretch}{0.95}
\begin{tabular}{@{}lccc@{}}
\toprule
\multirow{2}{*}{\bf Method}
& \bf Cam. & \bf Pixel & \bf 3D \\
& \bf Geo. & \bf Geo. & \bf Recon. \\
\midrule
FG$^2$ \citep{xia2025fg} & \cmark & \xmark & \xmark \\
Loc$^2$ \citep{xia2025loc} & \cmark & \cmark & \xmark \\
\midrule
MASt3R \citep{leroy2024grounding} & \xmark & \xmark & \cmark \\
VGGT \citep{wang2025vggt} & \xmark & \xmark & \cmark \\
A-MD \citep{vuong2025aerialmegadepth} & \xmark & \xmark & \cmark \\
\midrule
\textbf{X-VGGT} & \cmark & \cmark & \cmark \\
\bottomrule
\end{tabular}
\vspace{-2.0\baselineskip}
\end{wraptable}

We compare X-VGGT against cross-view geo-localization methods, multi-view 3D reconstruction models, and a baseline that aligns reconstructed geometry using predicted camera geo-locations. All methods are evaluated using the same scene inputs: 16 ground images and one georeferenced satellite image. Table~\ref{tab:baseline_prediction_support} summarizes the native prediction capabilities of the individual models.

\textbf{Cross-view geo-localization.}
FG$^2$~\citep{xia2025fg} and Loc$^2$~\citep{xia2025loc} estimate a ground camera's planar location and orientation within a satellite image. Each ground image is processed independently with the scene's satellite image. We report the north-up setting, without a ground-truth heading prior, and aggregate predictions over the ground views. For Loc$^2$, dense geo-registration is obtained by projecting its predicted depth using the estimated camera pose.

\textbf{3D reconstruction baselines.}
We evaluate MASt3R~\citep{leroy2024grounding}, VGGT~\citep{wang2025vggt}, and Aerial-MegaDepth~\citep{vuong2025aerialmegadepth} using both the ground and satellite images. The satellite is placed at index 0 as the reference view. VGGT jointly processes all 17 images, while MASt3R and Aerial-MegaDepth use a satellite-centered star graph pairing the satellite with each ground image, followed by sparse global alignment. Since these models only support perspective views, we predict a virtual perspective camera for the satellite reference, defined by it's intrinsic \(\hat K_{\mathrm{sat}}\) and extrinsic \([\hat R_{\mathrm{sat}}\mid\hat{\mathbf t}_{\mathrm{sat}}]\) parameters. We obtain satellite coordinates for each reconstructed camera center or dense point \(\mathbf X\) using
\[
    \hat{\mathbf u}(\mathbf X)
    =
    \mathcal N\!\left[
    \pi\!\left(
    \hat K_{\mathrm{sat}}
    \bigl(\hat R_{\mathrm{sat}}\mathbf X+\hat{\mathbf t}_{\mathrm{sat}}\bigr)
    \right)
    \right],
\]
where \(\pi\) denotes perspective projection and \(\mathcal N\) converts image coordinates to normalized satellite coordinates. No ground-truth pose, scale, or Sim(3) alignment is applied when computing geographic predictions for these baselines.

\textbf{Geo-localization-aligned reconstruction.}
Additionally, we evaluate a stronger baseline by aligning the 3D reconstruction and relative camera poses predicted by VGGT with the camera geo-locations independently predicted by Loc$^2$. VGGT reconstructs the 16 ground views without satellite, while Loc$^2$ independently predicts each ground camera's location in the satellite image. We estimate a common gravity direction from VGGT's predicted camera orientations and project the reconstruction into the ground plane. We then use RANSAC~\cite{fischler1981random} to fit a planar similarity transform between the projected camera centers and Loc$^2$'s predicted geo-locations. This defines a gravity-constrained Sim(3) transform, with zero vertical translation, that is applied consistently to all reconstructed cameras and dense 3D points. The alignment's yaw is inferred from camera-center correspondences, without using Loc$^2$'s per-frame yaw predictions.

\textbf{Evaluation policy.}
We convert normalized satellite-coordinate errors to meters using the known satellite resolution. Predicted coordinates are not clipped to the satellite boundary, which means that incorrect predictions can produce errors exceeding the 150\,m satellite extent. 

\subsection{Evaluation Metrics}
\label{sec:eval_metrics}

\textbf{Camera geo-localization} measures whether the predicted camera centers and orientation are correctly grounded in the satellite image. Since X-VGGT predicts camera locations in normalized satellite coordinates \(\hat{\mathbf{u}}_i,\mathbf{u}_i\in[0,1]^2\), we define the location error as \(e_i^{\mathrm{cam}}
    =
    \left\|
    (\hat{\mathbf{u}}_i-\mathbf{u}_i)
    \right\|_2
    \rho_{\mathrm{sat}},
\)
where \(\rho_{\mathrm{sat}}=150\) is the satellite scale in meters. Orientation error is computed by projecting each camera's principal ray into the satellite plane and measuring the angle between the predicted and ground-truth 2D directions.

\textbf{Pixel geo-registration} measures the distance between the ground truth and predicted pixel geo-locations. We compute this metric by projecting the image pixels into 3D space, and then into the satellite plane using the ground truth georeferencing transform. We report the location error between predicted and ground-truth satellite coordinates over valid pixels whose ground-truth projections lie inside the satellite image.

\textbf{3D reconstruction} evaluates the local multi-view geometry independently of satellite alignment. We align predicted camera poses to the metric ground-truth cameras with a similarity transform, then compute pairwise relative pose errors. We report relative rotation accuracy (RRA), relative translation accuracy (RTA), and their joint AUC at thresholds up to \(30^\circ\).

\begin{figure}[t]
  \centering
  \includegraphics[width=\textwidth]{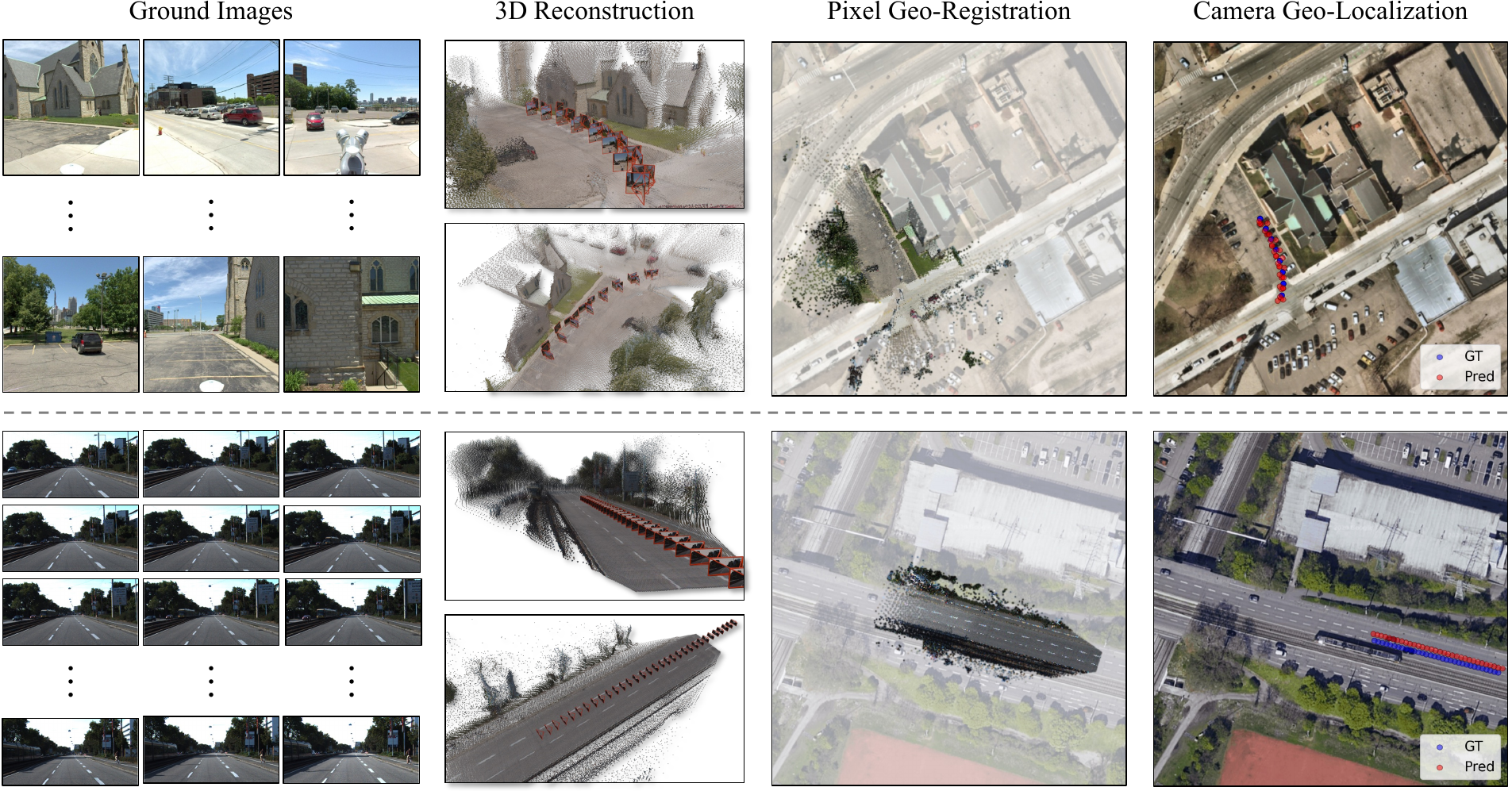}
\caption{\textbf{Qualitative results.}
We show two sample predictions from the Metropolis (top) and KITTI (bottom) datasets using X-VGGT. From left to right: input ground views, predicted 3D reconstruction, pixel-level geo-registration, and camera geo-localization. Our method jointly reconstructs the ground-view scene while projecting both image pixels and camera centers into the satellite coordinate frame.}
  \label{fig:qualitative}
\end{figure}

\subsection{Experimental Results}

\textbf{Camera Geo-Localization.} 
Table~\ref{tab:results_cam_geoloc} compares camera geo-localization across four datasets. X-VGGT achieves the lowest camera-center and orientation errors among the individual localization and reconstruction baselines, with average errors of \(22.38\)~m and \(30.93^\circ\). Cross-view localization methods remain competitive for predicting the camera-centers but produce large orientation errors, whereas 3D reconstruction baselines sometimes estimate orientation more accurately but have substantially larger localization errors. The largest localization gain occurs on Metropolis, where X-VGGT generalizes to an unseen multi-directional camera setup. Compared with geo-localization-aligned 3D reconstruction baseline (VGGT + Loc$^2$), our method reduces localization error by \(6.84\)~m on KITTI and \(22.71\)~m on Metropolis, while reducing orientation error from \(81.16^\circ\) to \(22.50^\circ\) and from \(89.19^\circ\) to \(17.59^\circ\), respectively. Overall, these results show that neither 2D cross-view localization, local 3D reconstruction or their combination are sufficient; robust georeferenced reconstruction requires jointly estimating scene geometry and its alignment to the satellite frame.

\begin{table}[t]
\centering
\caption{\textbf{Camera geo-localization.}
Mean camera-center localization error in meters and orientation error in degrees across four cross-view datasets. KITTI is evaluated on a geographically disjoint test split, while Metropolis, WRIVA, and ACC-NVS1 are used for zero-shot evaluation. VGGT + Loc$^2$ denotes geo-localization-aligned reconstruction, which aligns VGGT's reconstructed camera centers to locations independently predicted by Loc$^2$ using a shared Sim(3) transform. Dashes indicate results not reported.}
\label{tab:results_cam_geoloc}
\setlength{\tabcolsep}{4.5pt}
\renewcommand{\arraystretch}{1.08}
\resizebox{\textwidth}{!}{
\begin{tabular}{c|cccccccc}
\toprule
\multirow{2}{*}{\bf Method}
& \multicolumn{2}{c}{\bf KITTI}
& \multicolumn{2}{c}{\bf Metropolis}
& \multicolumn{2}{c}{\bf WRIVA}
& \multicolumn{2}{c}{\bf ACC-NVS1} \\
\cmidrule(lr){2-3}
\cmidrule(lr){4-5}
\cmidrule(lr){6-7}
\cmidrule(lr){8-9}
& Loc. (m) & Orien. ($^\circ$)
  & Loc. (m) & Orien. ($^\circ$)
  & Loc. (m) & Orien. ($^\circ$)
  & Loc. (m) & Orien. ($^\circ$) \\
\midrule
\midrule
FG$^2$ \citep{xia2025fg} & 39.81 & 89.00 & 40.26 & 89.73 & 36.47 & 85.80 & 36.51 & 77.58 \\
\addlinespace[2pt]
Loc$^2$ \cite{xia2025loc} & 32.00 & 69.99 & 37.61 & 84.09 & 36.30 & 76.16 & 39.96 & 80.81 \\
\addlinespace[2pt]
\midrule
MASt3R \citep{leroy2024grounding} & 1103.32 & 62.22 & 1028.05 & 79.83 & 1446.36 & 88.89 & 952.52 & 86.13 \\
VGGT \citep{wang2025vggt} & 130.05 & 32.17 & 408.13 & 60.77 & 455.36 & 63.00 & 168.45 & 77.54 \\
A-MD \citep{vuong2025aerialmegadepth} & 104.95 & 24.68 & 154.35 & 75.82 & 412.00 & 79.24 & 190.72 & 74.84 \\
\midrule
VGGT + Loc$^2$ & 31.46 & 81.16 & 37.06 & 89.19 & 35.16 & 87.41 & 37.07 & 89.41 \\
\midrule
\rowcolor{gray!15}
\textbf{X-VGGT (Ours)}
& \textbf{24.62} & \textbf{22.50}
& \textbf{14.35} & \textbf{17.59}
& \textbf{26.05} & \textbf{40.64}
& \textbf{24.49} & \textbf{42.99} \\
\bottomrule
\end{tabular}
}
\end{table}

\textbf{Pixel-Wise Geo-Registration.} 
Table~\ref{tab:results_pixel_geoloc} evaluates dense pixel geo-registration on KITTI and Metropolis, the two datasets with depth annotations. X-VGGT achieves the lowest errors, improving over the strongest baseline by \(18.44\)~m and \(32.70\)~m, respectively. The closest competitors are Loc$^2$ on KITTI and VGGT + Loc$^2$ on Metropolis. Compared directly with VGGT + Loc$^2$, X-VGGT reduces error from \(178.45\)~m to \(25.52\)~m on KITTI and from \(50.54\)~m to \(17.84\)~m on Metropolis. Although geo-localization-aligned reconstruction improves upon standalone Loc$^2$ on Metropolis, it substantially worsens dense registration on KITTI despite slightly improving camera-center localization. This indicates that fitting reconstructed camera centers to predicted geographic locations does not by itself ensure accurate geo-registration of the dense point maps, supporting joint estimation of scene geometry and satellite alignment.

\textbf{Local 3D Reconstruction.}
Table~\ref{tab:results_3d} shows that X-VGGT remains competitive on local 3D reconstruction, despite being trained to also estimate satellite alignment. It slightly improves over VGGT in most settings, likely because our model is initialized from the VGGT checkpoint while being further adapted to ground-level multi-view scenes. This indicates that adding georeferencing supervision does not degrade local 3D reconstruction quality.

\textbf{Qualitative Results.}
Figure~\ref{fig:qualitative} provides qualitative examples of X-VGGT predictions. The results show that our method produces coherent local 3D reconstructions while accurately projecting both dense points and camera centers into the satellite frame.

\begin{table}[t]
\centering
\begin{minipage}[t]{0.44\textwidth}
\centering
\caption{\textbf{Pixel-wise geo-registration on datasets with depth.}
Mean geo-localization error (meters) over ground-image pixels for KITTI and Mapillary Metropolis.}
\label{tab:results_pixel_geoloc}
\vspace{2pt}
\setlength{\tabcolsep}{6pt}
\renewcommand{\arraystretch}{1.05}
\resizebox{0.88\linewidth}{!}{
\begin{tabular}{c|cc}
\toprule
\bf Method & \bf KITTI & \bf Metropolis \\
\midrule
\midrule
Loc$^2$ \cite{xia2025loc} & 43.96 & 88.96 \\
\midrule
MASt3R \citep{leroy2024grounding} & 900.85 & 1500.67 \\
VGGT \citep{wang2025vggt} & 146.35 & 640.16 \\
A-MD \citep{vuong2025aerialmegadepth} & 62.07 & 319.15 \\
\midrule
VGGT + Loc$^2$ & 178.45 & 50.54 \\
\midrule
\rowcolor{gray!15}
\bf X-VGGT (Ours) & \bf 25.52 & \bf 17.84 \\
\bottomrule
\end{tabular}
}
\end{minipage}
\hfill
\begin{minipage}[t]{0.53\textwidth}
\centering
\caption{\textbf{Local 3D reconstruction.}
AUC of joint RRA and RTA up to \(30^\circ\). X-VGGT preserves the strong local reconstruction behavior of VGGT while adding georeferencing capabilities.}
\label{tab:results_3d}
\vspace{2pt}
\setlength{\tabcolsep}{3pt}
\renewcommand{\arraystretch}{1.05}
\resizebox{\linewidth}{!}{
\begin{tabular}{c|cccc}
\toprule
\bf Method & \bf KITTI & \bf Metropolis & \bf WRIVA & \bf ACC-NVS1 \\
\midrule
\midrule
MASt3R \citep{leroy2024grounding} & 6.64 & 5.05 & 0.03 & 6.25 \\
VGGT \citep{wang2025vggt} & 83.29 & \bf 55.31 & 16.90 & 67.96 \\
A-MD \citep{vuong2025aerialmegadepth} & 21.58 & 9.20 & 4.13 & 8.55 \\
\midrule
\rowcolor{gray!15}
\bf X-VGGT (Ours) & \bf 83.90 & 51.34 & \bf 21.17 & \bf 72.91 \\
\bottomrule
\end{tabular}
}
\end{minipage}
\end{table}

\FloatBarrier

\begin{figure}[t]
\centering
\begin{minipage}[t]{0.52\textwidth}
\vspace{0pt}
\centering
\captionof{table}{\textbf{Effect of georeferencing losses on KITTI \cite{geiger2013vision}.}
Adding dense point supervision and camera supervision improves both camera-level and pixel-level geolocation.}
\label{tab:loss_ablation}
\vspace{2pt}
\setlength{\tabcolsep}{6pt}
\renewcommand{\arraystretch}{1.08}
\begin{tabular}{cccc}
\toprule
\multirow{2}{*}{Losses}
& \multicolumn{2}{c}{Cam. Pose}
& 3D Points \\
\cmidrule(lr){2-3}
\cmidrule(lr){4-4}
& Loc. (m) & Orien. ($^\circ$) & Loc. (m) \\
\midrule
$\mathcal{L}_{\phi}$ & 30.35 & 32.87 & 33.17 \\
+ $\mathcal{L}_{\mathrm{dense}}$ & 27.95 & 34.49 & 30.91 \\
+ $\mathcal{L}_{\mathrm{center}}$ & \textbf{27.24} & \textbf{30.73} & \textbf{29.45} \\
\bottomrule
\end{tabular}
\end{minipage}
\hfill
\begin{minipage}[t]{0.42\textwidth}
\vspace{0pt}
\centering
\includegraphics[width=\linewidth]{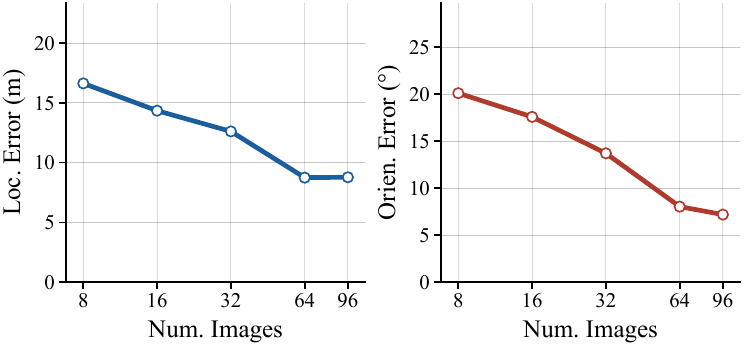}
\captionof{figure}{\textbf{Effect of the number of ground frames on Metropolis.}
Additional ground views reduce both camera localization and orientation errors.}
\label{fig:num_frames_ablation}
\end{minipage}
\end{figure}

\textbf{Ablations on the Effect of Georeferencing Losses.}
Table~\ref{tab:loss_ablation} ablates the proposed georeferencing losses on the KITTI dataset. Supervising only the georeferencing parameters provides a reasonable initial alignment, but adding dense projection supervision reduces the pixel-wise geo-registration error and improves camera localization. Adding camera-center supervision further improves the alignment, yielding the lowest errors across all reported metrics. These results show that the parameter, dense, and camera-center losses provide complementary signals for reducing both camera geo-localization and pixel-wise geo-registration errors.

\textbf{Ablations on the Effect of the Number of Ground Images.}
Figure~\ref{fig:num_frames_ablation} evaluates how performance changes with the number of input ground views on Mapillary Metropolis. Both localization and orientation errors decrease as more frames are provided, showing that X-VGGT effectively uses additional multi-view context to estimate a more accurate scene-level satellite alignment.

\section{Conclusion}

We introduced X-VGGT, a feed-forward model for georeferenced cross-view 3D reconstruction from ground and satellite images. X-VGGT jointly predicts ground-view cameras, dense 3D point maps, and a scene-level georeferencing transform that aligns the reconstruction with an orthorectified satellite image, enabling both camera geo-localization and dense pixel-wise geo-registration. We also proposed training and evaluation protocols covering ordered and unordered ground-view inputs across multiple datasets and sensor configurations. Experiments show that X-VGGT outperforms cross-view localization and 3D reconstruction baselines while preserving strong local 3D reconstruction quality, suggesting that robust georeferenced reconstruction requires coupling local multi-view geometry with explicit satellite-frame alignment.

\section*{Acknowledgments}
Supported by Intelligence Advanced Research Projects Activity (IARPA) via Department of Interior/Interior Business
Center (DOI/IBC) contract number 140D0423C0074. The
U.S. Government is authorized to reproduce and distribute
reprints for Governmental purposes notwithstanding any
copyright annotation thereon. Disclaimer: The views
and conclusions contained herein are those of the authors
and should not be interpreted as necessarily representing
the official policies or endorsements, either expressed or
implied, of IARPA, DOI/IBC, or the U.S. Government.

{
    \small
    \bibliographystyle{ieeenat_fullname}
    \bibliography{main}

@String(CVPR= {IEEE Conf. Comput. Vis. Pattern Recog.})

@String(ICCV= {Int. Conf. Comput. Vis.})

@String(ECCV= {Eur. Conf. Comput. Vis.})

@String(AAAI = {AAAI})

@String(CVPR  = {CVPR})

@String(ICCV  = {ICCV})

@String(ECCV  = {ECCV})

@inproceedings{leroy2024grounding,
  title={Grounding image matching in 3d with mast3r},
  author={Leroy, Vincent and Cabon, Yohann and Revaud, J{\'e}r{\^o}me},
  booktitle={European conference on computer vision},
  pages={71--91},
  year={2024},
  organization={Springer}
}

@article{wang2025pi,
  title={{$\pi^3$}: Permutation-Equivariant Visual Geometry Learning},
  author={Wang, Yifan and Zhou, Jianjun and Zhu, Haoyi and Chang, Wenzheng and Zhou, Yang and Li, Zizun and Chen, Junyi and Pang, Jiangmiao and Shen, Chunhua and He, Tong},
  journal={arXiv preprint arXiv:2507.13347},
  year={2025}
}

@article{lin2025depth,
  title={Depth anything 3: Recovering the visual space from any views},
  author={Lin, Haotong and Chen, Sili and Liew, Junhao and Chen, Donny Y and Li, Zhenyu and Shi, Guang and Feng, Jiashi and Kang, Bingyi},
  journal={arXiv preprint arXiv:2511.10647},
  year={2025}
}

@article{wang2026sky2ground,
  title={Sky2Ground: A Benchmark for Site Modeling under Varying Altitude},
  author={Wang, Zengyan and Mitra, Sirshapan and Modi, Rajat and Lim, Grace and Rawat, Yogesh},
  journal={arXiv preprint arXiv:2603.13740},
  year={2026}
}

@inproceedings{vuong2025aerialmegadepth,
  title={Aerialmegadepth: Learning aerial-ground reconstruction and view synthesis},
  author={Vuong, Khiem and Ghosh, Anurag and Ramanan, Deva and Narasimhan, Srinivasa and Tulsiani, Shubham},
  booktitle={Proceedings of the IEEE/CVF Conference on Computer Vision and Pattern Recognition},
  pages={21674--21684},
  year={2025}
}

@inproceedings{wang2024dust3r,
  title={Dust3r: Geometric 3d vision made easy},
  author={Wang, Shuzhe and Leroy, Vincent and Cabon, Yohann and Chidlovskii, Boris and Revaud, Jerome},
  booktitle={Proceedings of the IEEE/CVF conference on computer vision and pattern recognition},
  pages={20697--20709},
  year={2024}
}

@inproceedings{wang2025vggt,
  title={Vggt: Visual geometry grounded transformer},
  author={Wang, Jianyuan and Chen, Minghao and Karaev, Nikita and Vedaldi, Andrea and Rupprecht, Christian and Novotny, David},
  booktitle={Proceedings of the Computer Vision and Pattern Recognition Conference},
  pages={5294--5306},
  year={2025}
}

@InProceedings{wang2025cut3r,
    author    = {Wang, Qianqian and Zhang, Yifei and Holynski, Aleksander and Efros, Alexei A. and Kanazawa, Angjoo},
    title     = {Continuous 3D Perception Model with Persistent State},
    booktitle = {Proceedings of the IEEE/CVF Conference on Computer Vision and Pattern Recognition (CVPR)},
    month     = {June},
    year      = {2025},
    pages     = {10510-10522}
}

@inproceedings{ICLR2025_ce1710ff,
 author = {Zhang, Junyi and Herrmann, Charles and Hur, Junhwa and Jampani, Varun and darrell, trevor and Cole, Forrester and Sun, Deqing and Yang, Ming-Hsuan},
 booktitle = {International Conference on Learning Representations},
 editor = {Y. Yue and A. Garg and N. Peng and F. Sha and R. Yu},
 pages = {82863--82886},
 title = {MonST3R: A Simple Approach for Estimating Geometry in the Presence of Motion},
 url = {https://proceedings.iclr.cc/paper_files/paper/2025/file/ce1710ff5b66f89b0d707bf1ec2850bf-Paper-Conference.pdf},
 volume = {2025},
 year = {2025}
}

@InProceedings{Yang_2025_CVPR,
    author    = {Yang, Jianing and Sax, Alexander and Liang, Kevin J. and Henaff, Mikael and Tang, Hao and Cao, Ang and Chai, Joyce and Meier, Franziska and Feiszli, Matt},
    title     = {Fast3R: Towards 3D Reconstruction of 1000+ Images in One Forward Pass},
    booktitle = {Proceedings of the IEEE/CVF Conference on Computer Vision and Pattern Recognition (CVPR)},
    month     = {June},
    year      = {2025},
    pages     = {21924-21935}
}

@InProceedings{Cabon_2025_CVPR,
    author    = {Cabon, Yohann and Stoffl, Lucas and Antsfeld, Leonid and Csurka, Gabriela and Chidlovskii, Boris and Revaud, Jerome and Leroy, Vincent},
    title     = {MUSt3R: Multi-view Network for Stereo 3D Reconstruction},
    booktitle = {Proceedings of the IEEE/CVF Conference on Computer Vision and Pattern Recognition (CVPR)},
    month     = {June},
    year      = {2025},
    pages     = {1050-1060}
}

@InProceedings{Zhang_2025_CVPR,
    author    = {Zhang, Shangzhan and Wang, Jianyuan and Xu, Yinghao and Xue, Nan and Rupprecht, Christian and Zhou, Xiaowei and Shen, Yujun and Wetzstein, Gordon},
    title     = {FLARE: Feed-forward Geometry, Appearance and Camera Estimation from Uncalibrated Sparse Views},
    booktitle = {Proceedings of the IEEE/CVF Conference on Computer Vision and Pattern Recognition (CVPR)},
    month     = {June},
    year      = {2025},
    pages     = {21936-21947}
}

@inproceedings{keetha2026mapanything,
  title={{MapAnything}: Universal Feed-Forward Metric {3D} Reconstruction},
  author={Nikhil Keetha and Norman M\"{u}ller and Johannes Sch\"{o}nberger and Lorenzo Porzi and Yuchen Zhang and Tobias Fischer and Arno Knapitsch and Duncan Zauss and Ethan Weber and Nelson Antunes and Jonathon Luiten and Manuel Lopez-Antequera and Samuel Rota Bul\`{o} and Christian Richardt and Deva Ramanan and Sebastian Scherer and Peter Kontschieder},
  booktitle={International Conference on 3D Vision (3DV)},
  year={2026},
  organization={IEEE}
}

@inproceedings{wang2024vggsfm,
  title={Vggsfm: Visual geometry grounded deep structure from motion},
  author={Wang, Jianyuan and Karaev, Nikita and Rupprecht, Christian and Novotny, David},
  booktitle={Proceedings of the IEEE/CVF conference on computer vision and pattern recognition},
  pages={21686--21697},
  year={2024}
}

@InProceedings{Dong_2025_CVPR,
    author    = {Dong, Siyan and Wang, Shuzhe and Liu, Shaohui and Cai, Lulu and Fan, Qingnan and Kannala, Juho and Yang, Yanchao},
    title     = {Reloc3r: Large-Scale Training of Relative Camera Pose Regression for Generalizable, Fast, and Accurate Visual Localization},
    booktitle = {Proceedings of the IEEE/CVF Conference on Computer Vision and Pattern Recognition (CVPR)},
    month     = {June},
    year      = {2025},
    pages     = {16739-16752}
}

@inproceedings{huang2025c3po,
  title={C3Po: Cross-View Cross-Modality Correspondence by Pointmap Prediction},
  author={Huang, Kuan Wei and Li, Brandon and Hariharan, Bharath and Snavely, Noah},
  booktitle={The Thirty-ninth Annual Conference on Neural Information Processing Systems Datasets and Benchmarks Track},
  year={2025}
}

@article{geiger2013vision,
  title={Vision meets robotics: The kitti dataset},
  author={Geiger, Andreas and Lenz, Philip and Stiller, Christoph and Urtasun, Raquel},
  journal={The international journal of robotics research},
  volume={32},
  number={11},
  pages={1231--1237},
  year={2013},
  publisher={Sage Publications Sage UK: London, England}
}

@inproceedings{zhu2021vigor,
  title={Vigor: Cross-view image geo-localization beyond one-to-one retrieval},
  author={Zhu, Sijie and Yang, Taojiannan and Chen, Chen},
  booktitle={Proceedings of the IEEE/CVF Conference on Computer Vision and Pattern Recognition},
  pages={3640--3649},
  year={2021}
}

@misc{iarpa2024wriva,
  doi = {10.21227/cjk5-gf33},
  url = {https://dx.doi.org/10.21227/cjk5-gf33},
  author = {Myron Brown and Michael Chan and Michael Twardowski},
  publisher = {IEEE Dataport},
  title = {WRIVA Public Data},
  year = {2024} 
}

@article{sugg2025accnvs1,
  title={Accenture-nvs1: A novel view synthesis dataset},
  author={Sugg, Thomas and O'Brien, Kyle and Poudel, Lekh and Dumouchelle, Alex and Jou, Michelle and Bosch, Marc and Ramanan, Deva and Narasimhan, Srinivasa and Tulsiani, Shubham},
  journal={arXiv preprint arXiv:2503.18711},
  year={2025}
}

@inproceedings{shi2024weakly,
  title={Weakly-supervised camera localization by ground-to-satellite image registration},
  author={Shi, Yujiao and Li, Hongdong and Perincherry, Akhil and Vora, Ankit},
  booktitle={European Conference on Computer Vision},
  pages={39--57},
  year={2024},
  organization={Springer}
}

@article{wang2025bevsplat,
  title={BevSplat: Resolving height ambiguity via feature-based Gaussian primitives for weakly-supervised cross-view localization},
  author={Wang, Qiwei and Wu, Shaoxun and Shi, Yujiao},
  journal={arXiv preprint arXiv:2502.09080},
  year={2025}
}

@article{lehyeh2026geoflow,
  title={GeoFlow: Real-Time Fine-Grained Cross-View Geolocalization via Iterative Flow Prediction},
  author={Lehyeh, Ayesh Abu and Zhang, Xiaohan and Arrabi, Ahmad and Sultani, Waqas and Chen, Chen and Wshah, Safwan},
  journal={arXiv preprint arXiv:2603.21943},
  year={2026}
}

@article{wang2023fine,
  title={Fine-grained cross-view geo-localization using a correlation-aware homography estimator},
  author={Wang, Xiaolong and Xu, Runsen and Cui, Zhuofan and Wan, Zeyu and Zhang, Yu},
  journal={Advances in Neural Information Processing Systems},
  volume={36},
  pages={5301--5319},
  year={2023}
}

@article{song2023learning,
  title={Learning dense flow field for highly-accurate cross-view camera localization},
  author={Song, Zhenbo and Lu, Jianfeng and Shi, Yujiao and others},
  journal={Advances in Neural Information Processing Systems},
  volume={36},
  pages={70612--70625},
  year={2023}
}

@inproceedings{shi2022beyond,
  title={Beyond cross-view image retrieval: Highly accurate vehicle localization using satellite image},
  author={Shi, Yujiao and Li, Hongdong},
  booktitle={Proceedings of the IEEE/CVF Conference on Computer Vision and Pattern Recognition},
  pages={17010--17020},
  year={2022}
}

@inproceedings{wollam2026towards,
  title={Towards Unconstrained Cross-View Pose Estimation},
  author={Wollam, Alexander and Ashley, Kyle and Shugaev, Maxim and Arend, Oliver and Semenov, Ilya and Dashtestani, Hadis and Ravi, Sumved and Jacobs, Nathan},
  booktitle={Proceedings of the IEEE/CVF Winter Conference on Applications of Computer Vision},
  pages={8448--8457},
  year={2026}
}

@inproceedings{xia2025fg,
  title={FG\^{} 2: Fine-Grained Cross-View Localization by Fine-Grained Feature Matching},
  author={Xia, Zimin and Alahi, Alexandre},
  booktitle={Proceedings of the IEEE/CVF Conference on Computer Vision and Pattern Recognition},
  pages={6362--6372},
  year={2025}
}

@inproceedings{xia2025loc,
  title={Loc\^{} 2: Interpretable Cross-View Localization via Depth-Lifted Local Feature Matching},
  author={Xia, Zimin and Xu, Chenghao and Alahi, Alexandre},
  booktitle={The Fourteenth International Conference on Learning Representations},
  year={2025}
}

@inproceedings{liu2026pixel,
  title={Pixel-Wise Geo-Registration of Drone Images},
  author={Liu, Qingyang and Shatwell, David G and Kulkarni, Parth Parag and Shah, Mubarak},
  booktitle={European Conference on Computer Vision},
  pages={76--94},
  year={2026},
  organization={Springer}
}

@inproceedings{ye2024cross,
  title={Cross-view image geo-localization with Panorama-BEV Co-Retrieval Network},
  author={Ye, Junyan and Lv, Zhutao and Li, Weijia and Yu, Jinhua and Yang, Haote and Zhong, Huaping and He, Conghui},
  booktitle={European Conference on Computer Vision},
  pages={74--90},
  year={2024},
  organization={Springer}
}

@inproceedings{deuser2023sample4geo,
  title={Sample4geo: Hard negative sampling for cross-view geo-localisation},
  author={Deuser, Fabian and Habel, Konrad and Oswald, Norbert},
  booktitle={Proceedings of the IEEE/CVF International Conference on Computer Vision},
  pages={16847--16856},
  year={2023}
}

@inproceedings{li2024unleashing,
  title={Unleashing unlabeled data: A paradigm for cross-view geo-localization},
  author={Li, Guopeng and Qian, Ming and Xia, Gui-Song},
  booktitle={Proceedings of the IEEE/CVF Conference on Computer Vision and Pattern Recognition},
  pages={16719--16729},
  year={2024}
}

@inproceedings{zhu2022transgeo,
  title={Transgeo: Transformer is all you need for cross-view image geo-localization},
  author={Zhu, Sijie and Shah, Mubarak and Chen, Chen},
  booktitle={Proceedings of the IEEE/CVF Conference on Computer Vision and Pattern Recognition},
  pages={1162--1171},
  year={2022}
}

@inproceedings{vyas2022gama,
  title={Gama: Cross-view video geo-localization},
  author={Vyas, Shruti and Chen, Chen and Shah, Mubarak},
  booktitle={European Conference on Computer Vision},
  pages={440--456},
  year={2022},
  organization={Springer}
}

@inproceedings{pillai2024garet,
  title={Garet: cross-view video geolocalization with adapters and auto-regressive transformers},
  author={Pillai, Manu S and Rizve, Mamshad Nayeem and Shah, Mubarak},
  booktitle={European Conference on Computer Vision},
  pages={466--483},
  year={2024},
  organization={Springer}
}

@inproceedings{zhang2023cross,
  title={Cross-view image sequence geo-localization},
  author={Zhang, Xiaohan and Sultani, Waqas and Wshah, Safwan},
  booktitle={Proceedings of the IEEE/CVF Winter Conference on Applications of Computer Vision},
  pages={2914--2923},
  year={2023}
}

@inproceedings{torii2013visual,
  title={Visual place recognition with repetitive structures},
  author={Torii, Akihiko and Sivic, Josef and Pajdla, Tomas and Okutomi, Masatoshi},
  booktitle={Proceedings of the IEEE conference on computer vision and pattern recognition},
  pages={883--890},
  year={2013}
}

@inproceedings{arandjelovic2016netvlad,
  title={NetVLAD: CNN architecture for weakly supervised place recognition},
  author={Arandjelovic, Relja and Gronat, Petr and Torii, Akihiko and Pajdla, Tomas and Sivic, Josef},
  booktitle={Proceedings of the IEEE conference on computer vision and pattern recognition},
  pages={5297--5307},
  year={2016}
}

@inproceedings{berton2022rethinking,
  title={Rethinking visual geo-localization for large-scale applications},
  author={Berton, Gabriele and Masone, Carlo and Caputo, Barbara},
  booktitle={Proceedings of the IEEE/CVF Conference on Computer Vision and Pattern Recognition},
  pages={4878--4888},
  year={2022}
}

@article{ali2022gsv,
  title={Gsv-cities: Toward appropriate supervised visual place recognition},
  author={Ali-Bey, Amar and Chaib-Draa, Brahim and Giguere, Philippe},
  journal={Neurocomputing},
  volume={513},
  pages={194--203},
  year={2022},
  publisher={Elsevier}
}

@inproceedings{ali2023mixvpr,
  title={Mixvpr: Feature mixing for visual place recognition},
  author={Ali-Bey, Amar and Chaib-Draa, Brahim and Giguere, Philippe},
  booktitle={Proceedings of the IEEE/CVF winter conference on applications of computer vision},
  pages={2998--3007},
  year={2023}
}

@article{keetha2023anyloc,
  title={Anyloc: Towards universal visual place recognition},
  author={Keetha, Nikhil and Mishra, Avneesh and Karhade, Jay and Jatavallabhula, Krishna Murthy and Scherer, Sebastian and Krishna, Madhava and Garg, Sourav},
  journal={IEEE Robotics and Automation Letters},
  volume={9},
  number={2},
  pages={1286--1293},
  year={2023},
  publisher={IEEE}
}

@inproceedings{berton2023eigenplaces,
  title={Eigenplaces: Training viewpoint robust models for visual place recognition},
  author={Berton, Gabriele and Trivigno, Gabriele and Caputo, Barbara and Masone, Carlo},
  booktitle={Proceedings of the IEEE/CVF International Conference on Computer Vision},
  pages={11080--11090},
  year={2023}
}

@article{tzachor2024effovpr,
  title={Effovpr: Effective foundation model utilization for visual place recognition},
  author={Tzachor, Issar and Lerner, Boaz and Levy, Matan and Green, Michael and Shalev, Tal Berkovitz and Habib, Gavriel and Samuel, Dvir and Zailer, Noam Korngut and Shimshi, Or and Darshan, Nir and others},
  journal={arXiv preprint arXiv:2405.18065},
  year={2024}
}

@article{shatwell2026tiger,
  title={TIGeR: A Unified Framework for Time, Images and Geo-location Retrieval},
  author={Shatwell, David G and Swetha, Sirnam and Shah, Mubarak},
  journal={arXiv preprint arXiv:2603.24749},
  year={2026}
}

@article{astruc2026unigeoclip,
  title={UNIGEOCLIP: Unified Geospatial Contrastive Learning},
  author={Astruc, Guillaume and Trulls, Eduard and Hosang, Jan and Landrieu, Loic and Sarlin, Paul-Edouard},
  journal={arXiv preprint arXiv:2604.11668},
  year={2026}
}

@inproceedings{klemmer2025satclip,
  title={Satclip: Global, general-purpose location embeddings with satellite imagery},
  author={Klemmer, Konstantin and Rolf, Esther and Robinson, Caleb and Mackey, Lester and Ru{\ss}wurm, Marc},
  booktitle={Proceedings of the AAAI Conference on Artificial Intelligence},
  volume={39},
  number={4},
  pages={4347--4355},
  year={2025}
}

@inproceedings{clark2023we,
  title={Where we are and what we're looking at: Query based worldwide image geo-localization using hierarchies and scenes},
  author={Clark, Brandon and Kerrigan, Alec and Kulkarni, Parth Parag and Cepeda, Vicente Vivanco and Shah, Mubarak},
  booktitle={Proceedings of the IEEE/CVF Conference on Computer Vision and Pattern Recognition},
  pages={23182--23190},
  year={2023}
}

@inproceedings{haas2024pigeon,
  title={Pigeon: Predicting image geolocations},
  author={Haas, Lukas and Skreta, Michal and Alberti, Silas and Finn, Chelsea},
  booktitle={Proceedings of the IEEE/CVF Conference on Computer Vision and Pattern Recognition},
  pages={12893--12902},
  year={2024}
}

@article{vivanco2023geoclip,
  title={Geoclip: Clip-inspired alignment between locations and images for effective worldwide geo-localization},
  author={Vivanco Cepeda, Vicente and Nayak, Gaurav Kumar and Shah, Mubarak},
  journal={Advances in Neural Information Processing Systems},
  volume={36},
  pages={8690--8701},
  year={2023}
}

@inproceedings{seo2018cplanet,
  title={Cplanet: Enhancing image geolocalization by combinatorial partitioning of maps},
  author={Seo, Paul Hongsuck and Weyand, Tobias and Sim, Jack and Han, Bohyung},
  booktitle={Proceedings of the European Conference on Computer Vision (ECCV)},
  pages={536--551},
  year={2018}
}

@inproceedings{shatwell2025gt,
  title={Gt-loc: Unifying when and where in images through a joint embedding space},
  author={Shatwell, David G and Dave, Ishan Rajendrakumar and Swetha, Sirnam and Shah, Mubarak},
  booktitle={Proceedings of the IEEE/CVF International Conference on Computer Vision},
  pages={1--11},
  year={2025}
}

@InProceedings{Astruc_2024_CVPR,
    author    = {Astruc, Guillaume and Dufour, Nicolas and Siglidis, Ioannis and Aronssohn, Constantin and Bouia, Nacim and Fu, Stephanie and Loiseau, Romain and Nguyen, Van Nguyen and Raude, Charles and Vincent, Elliot and Xu, Lintao and Zhou, Hongyu and Landrieu, Loic},
    title     = {OpenStreetView-5M: The Many Roads to Global Visual Geolocation},
    booktitle = {Proceedings of the IEEE/CVF Conference on Computer Vision and Pattern Recognition (CVPR)},
    month     = {June},
    year      = {2024},
    pages     = {21967-21977}
}

@inproceedings{weyand2016planet,
  title={Planet-photo geolocation with convolutional neural networks},
  author={Weyand, Tobias and Kostrikov, Ilya and Philbin, James},
  booktitle={European conference on computer vision},
  pages={37--55},
  year={2016},
  organization={Springer}
}

@inproceedings{vo2017revisiting,
  title={Revisiting im2gps in the deep learning era},
  author={Vo, Nam and Jacobs, Nathan and Hays, James},
  booktitle={Proceedings of the IEEE international conference on computer vision},
  pages={2621--2630},
  year={2017}
}

@inproceedings{muller2018geolocation,
  title={Geolocation estimation of photos using a hierarchical model and scene classification},
  author={Muller-Budack, Eric and Pustu-Iren, Kader and Ewerth, Ralph},
  booktitle={Proceedings of the European conference on computer vision (ECCV)},
  pages={563--579},
  year={2018}
}

@inproceedings{pramanick2022world,
  title={Where in the world is this image? transformer-based geo-localization in the wild},
  author={Pramanick, Shraman and Nowara, Ewa M and Gleason, Joshua and Castillo, Carlos D and Chellappa, Rama},
  booktitle={European Conference on Computer Vision},
  pages={196--215},
  year={2022},
  organization={Springer}
}

@article{oquab2023dinov2,
  title={Dinov2: Learning robust visual features without supervision},
  author={Oquab, Maxime and Darcet, Timoth{\'e}e and Moutakanni, Th{\'e}o and Vo, Huy and Szafraniec, Marc and Khalidov, Vasil and Fernandez, Pierre and Haziza, Daniel and Massa, Francisco and El-Nouby, Alaaeldin and others},
  journal={arXiv preprint arXiv:2304.07193},
  year={2023}
}

@article{loshchilov2017decoupled,
  title={Decoupled weight decay regularization},
  author={Loshchilov, Ilya and Hutter, Frank},
  journal={arXiv preprint arXiv:1711.05101},
  year={2017}
}

@inproceedings{ranftl2021vision,
  title={Vision transformers for dense prediction},
  author={Ranftl, Ren{\'e} and Bochkovskiy, Alexey and Koltun, Vladlen},
  booktitle={2021 IEEE/CVF International Conference on Computer Vision (ICCV)},
  pages={12159--12168},
  year={2021},
  organization={IEEE}
}

@article{fischler1981random,
  title={Random sample consensus: a paradigm for model fitting with applications to image analysis and automated cartography},
  author={Fischler, Martin A and Bolles, Robert C},
  journal={Communications of the ACM},
  volume={24},
  number={6},
  pages={381--395},
  year={1981},
  publisher={ACM, New York, NY, USA}
}

@inproceedings{li2018megadepth,
  title={Megadepth: Learning single-view depth prediction from internet photos},
  author={Li, Zhengqi and Snavely, Noah},
  booktitle={2018 IEEE/CVF Conference on Computer Vision and Pattern Recognition},
  pages={2041--2050},
  year={2018},
  organization={IEEE}
}
}

\newpage
\appendix

\begin{center}
    {\Large\bfseries Appendix}
\end{center}
\vspace{1em}

\section{Limitations}
\label{app:limitations}

X-VGGT assumes access to an orthorectified, georeferenced satellite image with known spatial resolution. Performance may degrade when the satellite image is outdated, low resolution, occluded, or visually inconsistent with the ground views, since the model relies on cross-view evidence to align the local reconstruction to the satellite frame.

The model is also trained with satellite crops in a specific physical scale range, roughly \(150\)--\(250\) meters across depending on the dataset. Applying X-VGGT to substantially smaller or larger satellite regions may change the apparent scale and context seen by the model, and would likely require fine-tuning or adapting the crop protocol.

\section{Broader Impacts}
\label{app:broader_impacts}

This work studies georeferenced 3D reconstruction from ground-level and satellite images, unifying cross-view geo-localization with 3D reconstruction. We hope it can serve as an initial step toward treating these two tasks jointly. Potential positive impacts include improved mapping from ground images, localization in GPS-denied or GPS-degraded environments, disaster response using localized post-event images, infrastructure inspection, and robotics or autonomous navigation systems that need to connect local observations to a geographic map.

As with geo-localization methods more broadly, responsible use should consider privacy and data governance when applying the model to images of real-world places. The method should be used in accordance with dataset licenses, local regulations, and appropriate safeguards for images that may contain sensitive locations, private property, people, or vehicles. Since georeferencing errors can occur, predictions should be validated before use in safety-critical or decision-making settings.

\section{Additional Training Details}
\label{app:method_details}

\subsection{Training Satellite Images}

X-VGGT is trained on Sat-MegaDepth and SatKITTI. Sat-MegaDepth extends Aerial-MegaDepth with one orthorectified, georeferenced satellite mosaic per scene. It contains 121 scenes and 63,616 ground images, divided into 116 training scenes with 60,768 images and five validation scenes with 2,848 images. Table~\ref{tab:satmd_satellite_statistics} summarizes the physical coverage and native spatial resolution of the satellite mosaics across both splits.

\begin{figure}[t]
  \centering
  \includegraphics[width=\textwidth]{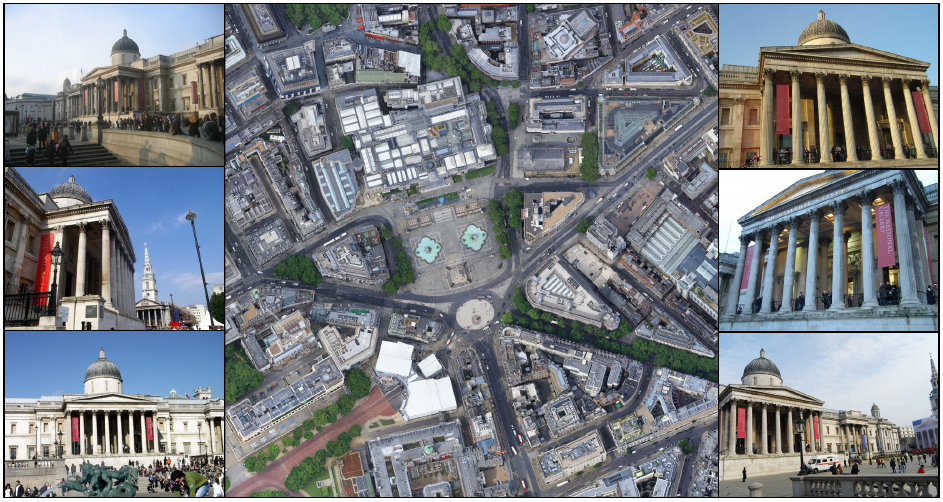}
\caption{\textbf{Example from the Sat-MegaDepth split.} The ground images at both sides are taken from the original MegaDepth~\cite{li2018megadepth} dataset. We obtain high-resolution orthorectified satellite images covering the location of all ground views from Google Maps.}
\label{fig:supp_satmd}
\end{figure}

Each training sample contains a set of ground images, a square satellite crop, camera calibration, ground-truth 3D points, and satellite-coordinate targets for the selected ground cameras. We sample satellite crops spanning 200--250~m per side for Sat-MegaDepth and 150--250~m for SatKITTI, then resize them to \(518\times518\) pixels. These crops provide different satellite contexts within the larger scene mosaics. Figure~\ref{fig:supp_satmd} shows an example from Sat-MegaDepth.

\begin{table}[ht]
\centering
\caption{\textbf{Sat-MegaDepth satellite imagery statistics.} Statistics are computed across the combined training and validation sets, comprising 121 satellite mosaics, before crop sampling or resizing. Physical dimensions are measured from the georeferenced image corners, and area is approximated as width \(\times\) height.}
\label{tab:satmd_satellite_statistics}
\setlength{\tabcolsep}{8pt}
\renewcommand{\arraystretch}{1.08}
\begin{tabular}{lrrrr}
\toprule
\bf Measurement & \bf Minimum & \bf Mean & \bf Median & \bf Maximum \\
\midrule
Width (m) & 502.2 & 748.7 & 617.8 & 1,649.6 \\
Height (m) & 502.2 & 760.9 & 627.4 & 2,018.6 \\
Area (km\(^2\)) & 0.263 & 0.628 & 0.390 & 2.206 \\
Native resolution (cm/pixel) & 7.42 & 10.41 & 10.46 & 12.99 \\
\bottomrule
\end{tabular}
\end{table}

\subsection{Coordinate Targets and Augmentation}

The satellite supervision uses normalized \([0,1]^2\) coordinates
with a vertical axis convention in which larger values correspond to upward
displacements in the satellite plane. Each dataloader returns normalized camera-center
targets \(\mathbf{u}^{\mathrm{cam}}_i\) and the satellite image scale in
meters. During training, the satellite image is
augmented with horizontal flips and random \(90^\circ\) rotations. The same
transform is applied to satellite-coordinate targets, the ENU world frame,
camera extrinsics, and point maps so that the georeferencing supervision
remains consistent after augmentation.

Before computing the losses, we apply the VGGT-style normalization of
camera extrinsics and point maps into the local reconstruction frame. Because this normalization changes the coordinate frame in which the
model predicts point maps and camera poses, we recompute the
ground-truth georeferencing parameters \(\phi^*\), dense satellite targets, and
valid dense projection masks after normalization. This recomputation keeps
\(\mathcal{G}_{\phi^*}\), camera-center targets, and dense point targets in the
same coordinate frame as the model predictions.

\subsection{Optimization}

The model is initialized from the pretrained VGGT checkpoint. New satellite
camera/register tokens and the Satellite Alignment Head (SAH) are optimized
from the start of training to learn the cross-view alignment needed for
georeferencing, while the camera and point heads retain their reconstruction
roles. The top four global attention blocks are unfrozen after \(10\%\) of
training, while the camera head, point head, and remaining default parameters
are unfrozen after \(40\%\) of training. We use separate warmup and cosine
schedules for each parameter group, with the largest learning rate for the SAH
and smaller learning rates for the pretrained backbone and 3D heads. Training
batches contain 4--32 ground images per scene; all images are resized to
\(518\times 518\) pixels, with zero-padding for rectangular aspect ratios. We use the AdamW optimizer~\cite{loshchilov2017decoupled} with weight decay \(0.05\), bfloat16
automatic mixed precision, gradient clipping, and train for 80,000 iterations on four H100 GPUs. Models trained for the georeferencing loss ablation in Table~\ref{tab:loss_ablation} use the same training schedule, but are trained for 10,000 iterations on two H100 GPUs.

\section{Additional Evaluation Details}
\label{app:evaluation_details}

\subsection{Evaluation datasets.}
\textbf{KITTI}~\citep{geiger2013vision} is an autonomous driving dataset collected around Karlsruhe, Germany, including urban roads, rural areas, and highways. The capture platform is equipped with four cameras, two grayscale and two RGB, a Velodyne LiDAR scanner, and a high-precision GPS/IMU system. We use the augmented benchmark of \citet{shi2022beyond}, which provides satellite images around the vehicle trajectory. For our evaluation, we use only the left RGB camera: the left and right stereo views are highly similar, and our goal is to evaluate sequence-level ground-satellite georeferencing rather than stereo reconstruction. We use LiDAR-derived depth maps for pixel-wise evaluation and report results on the challenging cross-area Test-2 split.

\textbf{Mapillary Metropolis} is a city-scale driving dataset collected in Chicago. It provides synchronized camera poses, georeferenced aerial imagery, panoramic street-level images, and four perspective images at each timestamp, corresponding to the front, back, left, and right directions relative to the vehicle. The dataset also includes multi-view-stereo reconstructions, MVS-derived depth maps for the perspective images, and LiDAR point clouds. Although LiDAR points are available, they are sparser than the MVS-derived depth maps for dense image evaluation, so we use the MVS depth maps when computing pixel-wise geo-registration metrics. We use Metropolis for zero-shot evaluation, since it differs from our training data in sensor layout, capture protocol, and scene distribution.

\textbf{WRIVA} is derived from the outdoor sites of the WRIVA Public Dataset~\citep{iarpa2024wriva}. Unlike KITTI and Metropolis, which are vehicle-centric driving datasets, WRIVA contains unstructured image collections from six U.S. sites with cameras scattered around each scene. The scenes have substantially different layouts from urban driving datasets, consisting largely of rural environments, open areas, and training-ground-style sites relevant to first-responder and site-modeling scenarios. WRIVA provides calibrated camera parameters and GPS annotations. Although the original dataset includes low-resolution panchromatic and multispectral satellite imagery, these modalities are outside the scope of our work, so we replace them with high-resolution orthorectified RGB satellite images from Google Maps. Since WRIVA does not provide ground-truth depth maps, we use it only for camera geo-localization and pose evaluation.

\textbf{ACC-NVS1}~\citep{sugg2025accnvs1} is a multi-view dataset collected for novel view synthesis across six real-world scenes in Austin, TX and Pittsburgh, PA. It contains approximately 148K images captured from both ground and airborne viewpoints, with calibrated camera intrinsics, extrinsics, and geolocation. In our evaluation, we use the ground-level image sequences and supplement them with high-resolution orthorectified RGB satellite images to support cross-view geo-localization. Since ACC-NVS1 does not provide ground-truth depth maps, we use it only for camera geo-localization and pose evaluation.

\subsection{Evaluation scene generation.}
We construct evaluation scenes using dataset-specific sampling protocols that reflect the structure of each dataset while keeping the evaluation format consistent across benchmarks.

\textbf{KITTI} scenes are generated from linear driving sequences. We sample contiguous or nearby frames from each sequence to form ordered ground-view sets, using only the left RGB camera. This produces sequence-like scenes where the model must estimate a shared satellite alignment from multiple views along the vehicle trajectory.

\textbf{Mapillary Metropolis} scenes are constructed from consecutive capture groups. Each capture group contains four perspective images aligned with the vehicle frame, corresponding to orthogonal viewing directions. We preserve this multi-directional structure when creating the scenes, which provides denser coverage around the vehicle than a single forward-facing driving sequence.

\textbf{WRIVA} contains unordered image collections rather than vehicle trajectories. We follow a VGGT-style batch construction strategy: for each scene, we choose a seed image and sample additional views with high pairwise pose similarity. This produces coherent local scenes while preserving the unordered nature of the dataset.

\textbf{ACC-NVS1} scenes are generated from ground-level image sequences. Similar to KITTI, we sample ordered frame sets from each sequence to evaluate camera geo-localization under trajectory-like motion, but on locations and capture conditions not seen during training.

\textbf{Evaluation Satellite Images.}
For every evaluation scene, we compute the centroid of the selected camera centers and extract a \(150\mathrm{m}\times150\mathrm{m}\) orthorectified satellite crop. Rather than placing the centroid exactly at the crop center, we randomly position it within the central \(100\mathrm{m}\times100\mathrm{m}\) region. This prevents the evaluation from assuming a perfectly centered satellite crop while ensuring that the selected cameras remain inside the image. For the main results, we sample up to \(1024\) scenes per dataset with \(16\) ground images per scene. For the ablation on the number of ground images, we construct larger scenes with up to \(96\) ground images.

\section{Metric Depth Estimation}
\label{app:metric_depth}

\begin{figure}[htbp]
  \centering
  \includegraphics[width=\textwidth]{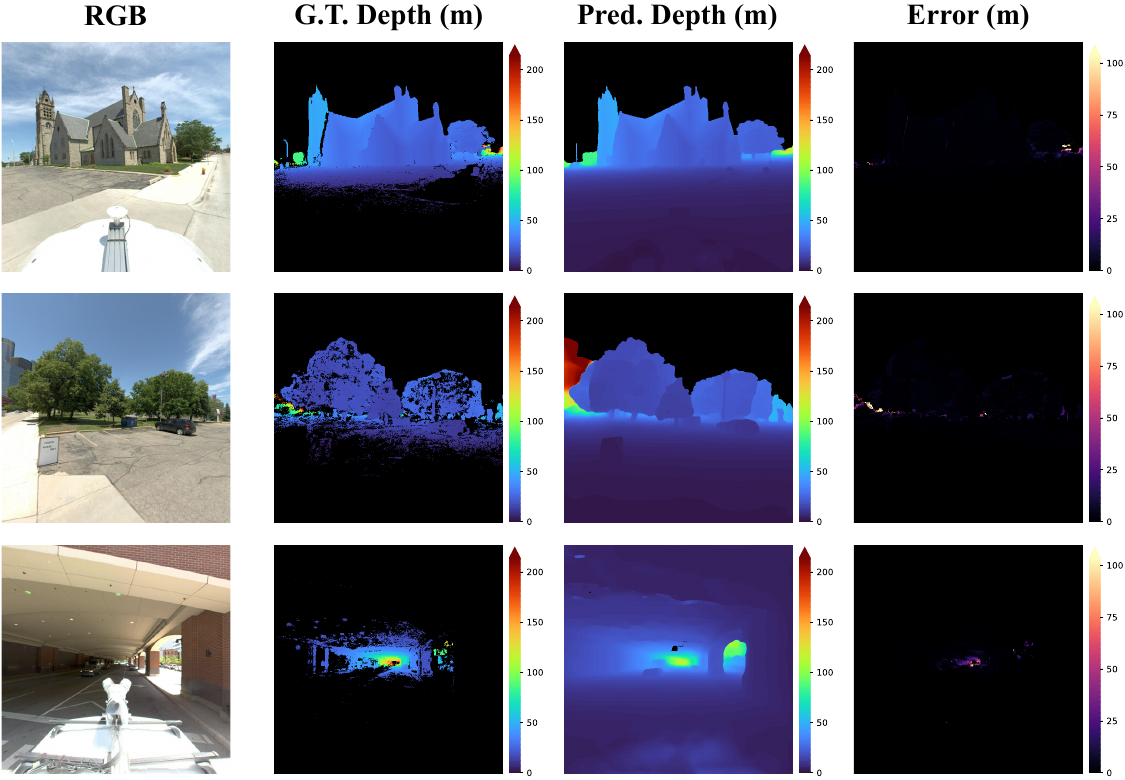}
\caption{\textbf{Qualitative metric depth predictions.} From left to right: input ground image, ground-truth depth, X-VGGT metric depth, and absolute depth error. Ground-truth and predicted depth maps share the same colormap and range across all examples; error maps use a separate shared scale. All values are in meters. Missing ground-truth depth is shown in black, and errors are computed only where ground-truth depth is available.}
\label{fig:metric_depth}
\end{figure}

The known ground sampling distance of the satellite image provides a metric reference for the reconstructed scene, allowing X-VGGT to resolve the scale ambiguity in the ground 3D reconstruction and generate metric depth maps. The Satellite Alignment Head predicts a log-scale \(\alpha\) that maps reconstruction units to normalized satellite coordinates. For a square satellite crop spanning \(\rho_{\mathrm{sat}}\) meters per side, the corresponding metric scale is \(\rho_{\mathrm{sat}}\exp(\alpha)\). We transform each predicted point into its ground camera's coordinate frame and extract its depth:
\begin{equation}
\hat d_i(p)
=
\rho_{\mathrm{sat}}\exp(\alpha)\,
\left[
\hat R_i\hat{\mathbf X}_i(p)+\hat{\mathbf t}_i
\right]_z,
\label{eq:metric_depth}
\end{equation}
where \(\hat{\mathbf X}_i(p)\) is the predicted 3D point at pixel \(p\), \([\hat R_i\mid\hat{\mathbf t}_i]\) is the predicted camera pose w.r.t. the reference camera, and \([\cdot]_z\) selects the camera-frame depth component. Figure~\ref{fig:metric_depth} compares the ground truth and predicted depth maps, showing that X-VGGT is able to successfully recover the metric scale from the georeferenced satellite image.

\FloatBarrier

\section{Effect of Ground--Satellite Interaction}
\label{app:ground_satellite_interaction}

To study the role of joint cross-view encoding, we compare X-VGGT with a late-fusion baseline that processes the ground views and satellite image in separate passes through a shared encoder. The ground views retain token-level interaction with one another, but ground and satellite tokens do not interact directly. The alignment head processes the two streams separately, pools the ground camera tokens, and combines the ground and satellite summaries with an MLP to predict the georeferencing transform. The camera and point heads receive only ground-view features.

Table~\ref{tab:late_fusion_ablation} shows that late fusion increases camera localization, orientation, and dense registration errors on both KITTI and Metropolis. For example, KITTI localization error rises from \(24.62\)~m to \(43.58\)~m, while orientation error rises from \(22.50^\circ\) to \(95.07^\circ\). These results support the value of jointly encoding ground and satellite views for estimating the scene-level alignment.

\begin{table}[h]
\centering
\caption{\textbf{Effect of ground--satellite interaction.} Mean camera localization error (Loc.), orientation error (Orien.), and pixel-wise geo-registration error (G.R.) for the late-fusion baseline and X-VGGT. Both methods use 16 ground images and one satellite image. Lower is better.}
\label{tab:late_fusion_ablation}
\setlength{\tabcolsep}{6pt}
\renewcommand{\arraystretch}{1.08}
\begin{tabular}{lcccccc}
\toprule
\multirow{2}{*}{\bf Method}
& \multicolumn{3}{c}{\bf KITTI}
& \multicolumn{3}{c}{\bf Metropolis} \\
\cmidrule(lr){2-4}
\cmidrule(lr){5-7}
& Loc. (m) & Orien. ($^\circ$) & G.R. (m)
& Loc. (m) & Orien. ($^\circ$) & G.R. (m) \\
\midrule
Late fusion & 43.58 & 95.07 & 44.21 & 41.96 & 87.26 & 54.81 \\
\textbf{X-VGGT} & \textbf{24.62} & \textbf{22.50} & \textbf{25.52} & \textbf{14.35} & \textbf{17.59} & \textbf{17.84} \\
\bottomrule
\end{tabular}
\end{table}

\section{Inference Efficiency}
\label{app:inference_efficiency}

We measure X-VGGT's inference latency and peak GPU memory as the number of ground images increases from 4 to 64. Experiments use an NVIDIA A100 80GB GPU, batch size of 1 scene, image resolution of \(518\times518\) pixels, and FP16 mixed precision. After 10 warm-up iterations, we report the median latency and peak PyTorch-allocated memory over 30 forward passes, synchronizing the GPU for timing. Latency excludes model loading, data loading, preprocessing, and CPU-to-GPU tensor transfer.

\begin{table}[h]
\centering
\caption{\textbf{Inference efficiency.} Inference latency and peak GPU memory of X-VGGT for different numbers of ground images. Every setting additionally includes one satellite image.}
\label{tab:inference_efficiency}
\setlength{\tabcolsep}{10pt}
\renewcommand{\arraystretch}{1.08}
\begin{tabular}{ccc}
\toprule
\bf Ground images & \bf Latency (s) & \bf Peak GPU memory (GiB) \\
\midrule
4 & 0.244 & 9.30 \\
8 & 0.470 & 11.50 \\
16 & 1.020 & 13.63 \\
32 & 2.579 & 17.90 \\
64 & 7.539 & 26.65 \\
\bottomrule
\end{tabular}
\end{table}

Table~\ref{tab:inference_efficiency} shows that the 16-ground-view setting used in the main experiments requires \(1.020\)~s and \(13.63\)~GiB of GPU memory. Increasing the number of views raises both computational cost and memory usage, with latency growing more rapidly at larger view counts. These measurements complement the view-count accuracy ablation in Figure~\ref{fig:num_frames_ablation} by quantifying the cost of incorporating additional views.

\section{Robustness to Satellite Configurations}
\label{app:satellite_robustness}

X-VGGT uses satellite imagery to place reconstructed cameras and 3D points in a geographic coordinate frame. We examine how this alignment is affected by the area shown in the satellite image, the amount of visual detail it contains, and errors in its geographic coordinates. On KITTI and Metropolis, we test these factors separately by changing the spatial dimentions, reducing the resolution, and introducing spatial offsets, while keeping each scene's 16 ground images fixed. To compare all settings on the same scenes, we require the cameras to remain visible within the smallest satellite crop and sufficient surrounding imagery to extract larger or shifted crops without padding. Specifically, all camera centers must lie at least 5~m inside the smallest crop, and the available imagery must accommodate the largest crop and shifts of up to 10~m. We select 512 scenes per dataset that satisfy these requirements using a fixed seed of 0, before inference and independently of model predictions.

These requirements exclude scenes with cameras spread over a large area or positioned near the satellite boundary. Our main evaluation instead uses an independent selection of 1,024 scenes per dataset without these restrictions. The two evaluations therefore contain different scenes, which must be considered when interpreting their results---for example, the unmodified satellite images yield a KITTI localization error of \(12.88\)~m here, compared with \(24.62\)~m in Table~\ref{tab:results_cam_geoloc}. To measure the effect of each change, we compare it against the unmodified satellite image using exactly the same 512 scenes.

\textbf{Satellite crop size.} We vary the width and height of the satellite crop by \(\pm25\%\) around the standard \(150\times150\)~m region, keeping the crop center and ground images fixed. Each crop is resized to \(518\times518\) pixels, and we update its geographic bounds and the corresponding evaluation targets. Table~\ref{tab:satellite_crop_robustness} shows that larger crops increase localization and geo-registration errors on both datasets. Smaller crops yield similar KITTI localization error and lower Metropolis errors, although KITTI orientation error increases.

\begin{table}[h]
\centering
\caption{\textbf{Robustness to satellite crop size.} Mean camera localization error (Loc.), orientation error (Orien.), and pixel-wise geo-registration error (G.R.) on the same 512 scenes per dataset. Crop size denotes the side length of the square satellite region; all crops are resized to \(518\times518\) pixels. Lower is better.}
\label{tab:satellite_crop_robustness}
\setlength{\tabcolsep}{6pt}
\renewcommand{\arraystretch}{1.08}
\begin{tabular}{lcccccc}
\toprule
\multirow{2}{*}{\bf Crop size}
& \multicolumn{3}{c}{\bf KITTI}
& \multicolumn{3}{c}{\bf Metropolis} \\
\cmidrule(lr){2-4}
\cmidrule(lr){5-7}
& Loc. (m) & Orien. ($^\circ$) & G.R. (m)
& Loc. (m) & Orien. ($^\circ$) & G.R. (m) \\
\midrule
112.5 m (\(0.75\times\)) & 12.78 & 17.55 & 13.85 & 9.36 & 12.99 & 11.81 \\
150 m (\(1\times\)) & 12.88 & 14.71 & 13.17 & 10.53 & 13.36 & 13.77 \\
187.5 m (\(1.25\times\)) & 17.06 & 15.42 & 15.78 & 13.27 & 13.89 & 17.65 \\
\bottomrule
\end{tabular}
\end{table}

\textbf{Satellite resolution.} We evaluate the robustness of X-VGGT to different satellite resolutions by downsampling the \(518\times518\) satellite image to \(259\times259\) or \(130\times130\) pixels, then resize it back to \(518\times518\) using bicubic interpolation. The ground images, geographic bounds, and evaluation targets remain fixed. Table~\ref{tab:satellite_resolution_robustness} shows increasing errors as satellite detail is reduced. At \(130\times130\) pixels, localization error rises from \(12.88\)~m to \(25.28\)~m on KITTI and from \(10.53\)~m to \(20.04\)~m on Metropolis, highlighting the importance of spatial detail for cross-view alignment.

\begin{table}[h]
\centering
\caption{\textbf{Robustness to satellite resolution.} Satellite images are downsampled to the listed resolution and then resized back to the model's \(518\times518\) input resolution, without changing the geographic region. Metrics and scenes are the same as in Table~\ref{tab:satellite_crop_robustness}. Lower is better.}
\label{tab:satellite_resolution_robustness}
\setlength{\tabcolsep}{6pt}
\renewcommand{\arraystretch}{1.08}
\begin{tabular}{lcccccc}
\toprule
\multirow{2}{*}{\bf Resolution}
& \multicolumn{3}{c}{\bf KITTI}
& \multicolumn{3}{c}{\bf Metropolis} \\
\cmidrule(lr){2-4}
\cmidrule(lr){5-7}
& Loc. (m) & Orien. ($^\circ$) & G.R. (m)
& Loc. (m) & Orien. ($^\circ$) & G.R. (m) \\
\midrule
\(518\times518\) ($1.00 \times$) & 12.88 & 14.71 & 13.17 & 10.53 & 13.36 & 13.77 \\
\(259\times259\) ($0.50 \times$) & 19.90 & 17.78 & 18.25 & 14.16 & 13.81 & 17.37 \\
\(130\times130\) ($0.25 \times$) & 25.28 & 24.68 & 23.88 & 20.04 & 22.49 & 25.57 \\
\bottomrule
\end{tabular}
\end{table}

\textbf{Satellite misalignment.} To simulate errors in satellite georeferencing, we randomly shift the satellite images by 5 or 10~m in any cardinal direction, while retaining the original geographic bounds and evaluation targets. This introduces a mismatch between the image content and its assigned geographic coordinates. Table~\ref{tab:satellite_misalignment_robustness} shows that these shifts primarily affect localization and geo-registration, while orientation errors remain comparatively stable. With a 10~m shift, localization error increases from \(12.88\)~m to \(16.74\)~m on KITTI and from \(10.53\)~m to \(14.80\)~m on Metropolis.

\begin{table}[h]
\centering
\caption{\textbf{Robustness to satellite misalignment.} The satellite image is shifted by the listed distance while its assigned geographic bounds and evaluation targets are held fixed. Shift directions are balanced across north, east, south, and west. Metrics and scenes are the same as in Table~\ref{tab:satellite_crop_robustness}. Lower is better.}
\label{tab:satellite_misalignment_robustness}
\setlength{\tabcolsep}{6pt}
\renewcommand{\arraystretch}{1.08}
\begin{tabular}{lcccccc}
\toprule
\multirow{2}{*}{\bf Image shift}
& \multicolumn{3}{c}{\bf KITTI}
& \multicolumn{3}{c}{\bf Metropolis} \\
\cmidrule(lr){2-4}
\cmidrule(lr){5-7}
& Loc. (m) & Orien. ($^\circ$) & G.R. (m)
& Loc. (m) & Orien. ($^\circ$) & G.R. (m) \\
\midrule
0 m & 12.88 & 14.71 & 13.17 & 10.53 & 13.36 & 13.77 \\
5 m & 14.27 & 14.80 & 14.50 & 11.96 & 13.08 & 15.09 \\
10 m & 16.74 & 15.36 & 17.05 & 14.80 & 13.29 & 17.86 \\
\bottomrule
\end{tabular}
\end{table}

\section{Oracle Alignment}
\label{app:oracle_alignment}

The geographic accuracy of X-VGGT depends on both the reconstructed geometry and its alignment with the satellite image. To examine how much error can be reduced by improving alignment, we compare the predicted alignment with an oracle obtained from ground-truth camera locations. For each scene, we fit a single Sim(3) transform, consisting of a rotation, translation, and uniform scale, between X-VGGT's 16 predicted camera centers and their ground-truth 3D locations. We apply this transform to all predicted cameras and dense points, then project them into the satellite plane using the ground-truth satellite transform.

Table~\ref{tab:oracle_alignment} shows that oracle alignment substantially reduces camera localization and orientation errors on both datasets. On KITTI, localization error decreases from \(24.62\)~m to \(0.09\)~m, while pixel-wise geo-registration error decreases from \(25.52\)~m to \(7.65\)~m, indicating considerable room to improve the predicted alignment. On Metropolis, camera localization improves from \(14.35\)~m to \(1.88\)~m and orientation error decreases from \(17.59^\circ\) to \(3.78^\circ\), but dense geo-registration improves only from \(17.84\)~m to \(16.84\)~m. Thus, accurately aligning camera centers does not by itself ensure accurate placement of the dense reconstruction.

\begin{table}[h]
\centering
\caption{\textbf{Oracle alignment comparison.} Mean camera localization error (Loc.), orientation error (Orien.), and pixel-wise geo-registration error (G.R.) for X-VGGT using its predicted alignment or a scene-level alignment fitted to ground-truth camera centers. Both settings use the same predicted local reconstruction. Lower is better.}
\label{tab:oracle_alignment}
\setlength{\tabcolsep}{5pt}
\renewcommand{\arraystretch}{1.08}
\begin{tabular}{lcccccc}
\toprule
\multirow{2}{*}{\bf Method}
& \multicolumn{3}{c}{\bf KITTI}
& \multicolumn{3}{c}{\bf Metropolis} \\
\cmidrule(lr){2-4}
\cmidrule(lr){5-7}
& Loc. (m) & Orien. ($^\circ$) & G.R. (m)
& Loc. (m) & Orien. ($^\circ$) & G.R. (m) \\
\midrule
X-VGGT & 24.62 & 22.50 & 25.52 & 14.35 & 17.59 & 17.84 \\
X-VGGT + oracle alignment & 0.09 & 6.49 & 7.65 & 1.88 & 3.78 & 16.84 \\
\bottomrule
\end{tabular}
\end{table}

\section{Additional Qualitative Examples}
\label{app:qualitative}

\begin{figure}[t]
  \centering
  \includegraphics[width=\textwidth]{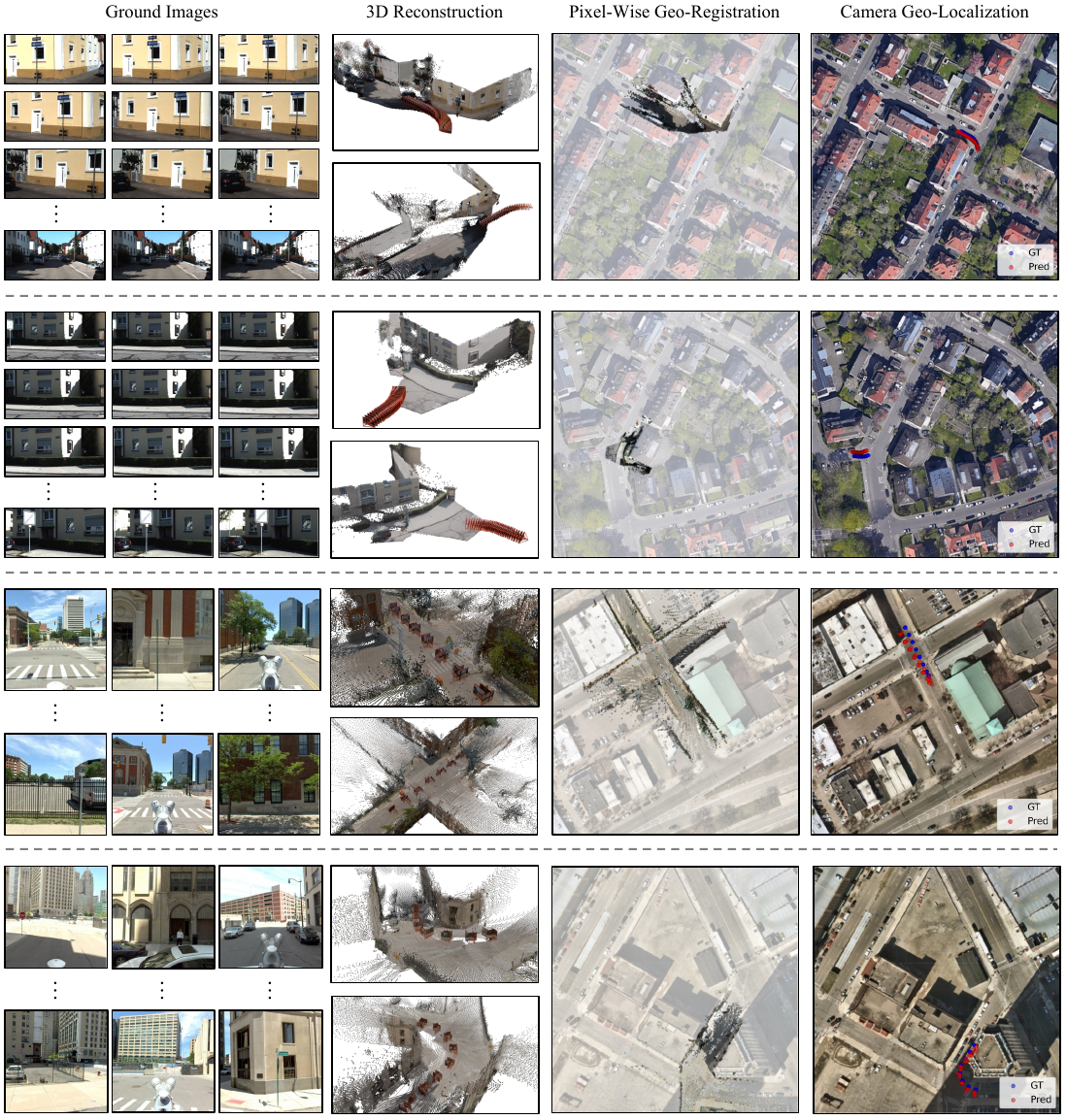}
\caption{\textbf{Additional qualitative predictions.} We show more example predictions from X-VGGT across scenes from KITTI (first two rows) and Mapillary Metropolis (last two rows).}
\label{fig:supp_qual}
\end{figure}

\begin{figure}[t]
  \centering
  \includegraphics[width=\textwidth]{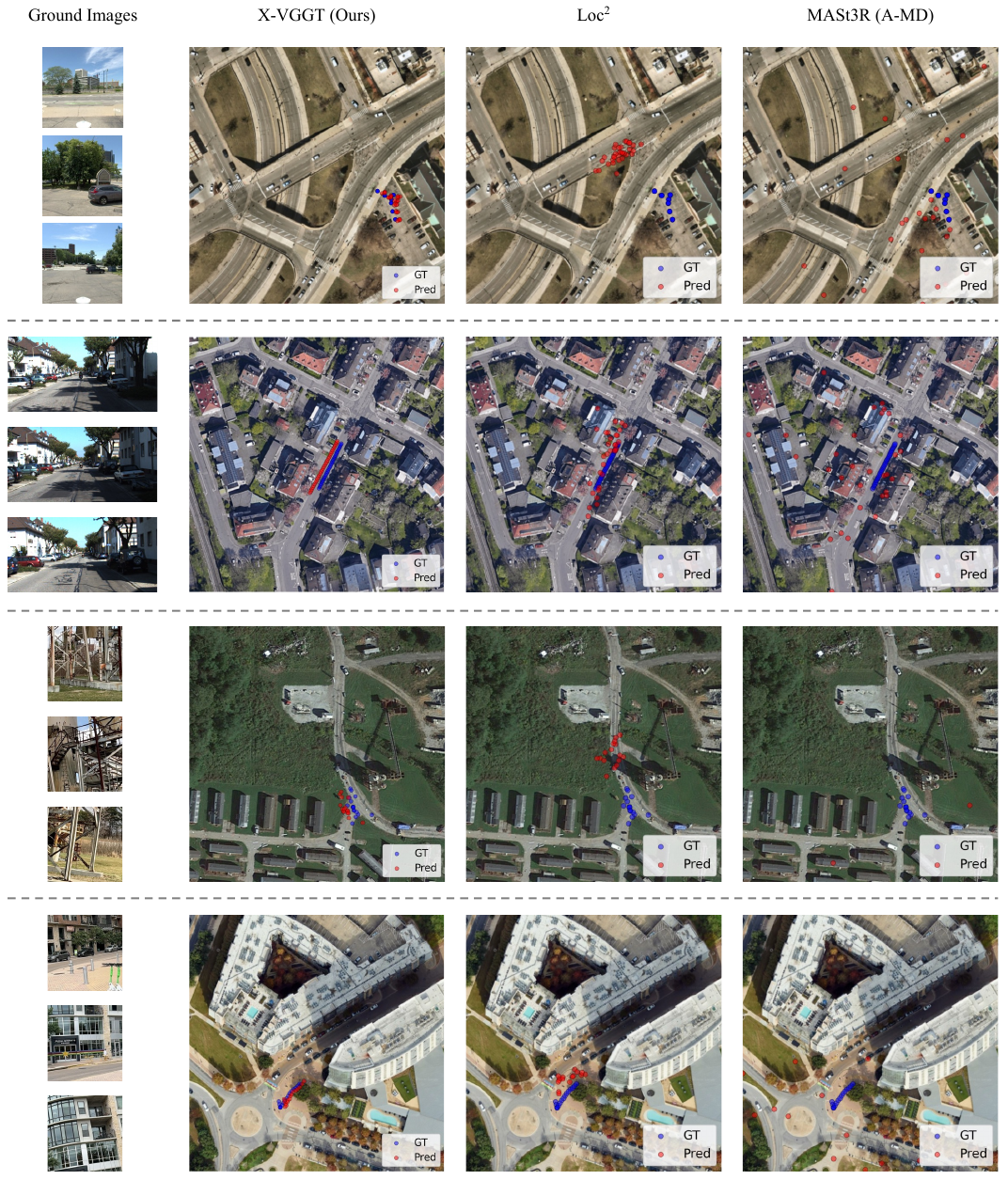}
\caption{\textbf{Qualitative comparison with baselines.} We compare X-VGGT against the baselines on camera geo-localization across the four evaluation datasets. From top to bottom: KITTI, Mapillary Metropolis, WRIVA, and ACC-NVS1.}
\label{fig:supp_qual_baselines}
\end{figure}

Figure~\ref{fig:supp_qual} shows additional X-VGGT prediction examples across evaluation scenes. Figure~\ref{fig:supp_qual_baselines} compares X-VGGT against baselines on camera geo-localization and shows that our model produces more accurate predictions across all datasets.

\section{Licenses and Existing Assets}
\label{app:licenses}

We use existing datasets and pretrained models, and cite the original sources throughout the paper. KITTI and Mapillary Metropolis are released under a Creative Commons Attribution-NonCommercial-ShareAlike license, and MegaDepth-derived data is released under a Creative Commons Attribution license. We use these assets for research purposes and include qualitative examples in accordance with the corresponding dataset licenses and terms. The paper is distributed under a license compatible with the reproduced third-party visual assets.

\end{document}